\documentclass[journal]{IEEEtran}

\usepackage[subfigure]{graphfig}

\usepackage{graphicx}
\usepackage{algorithm}
\usepackage{algorithmic}
\usepackage{mathrsfs} 
\usepackage{amsmath}
\usepackage{makecell}
\usepackage{cite}

\newcommand{\tabincell}[2]{\begin{tabular}{@{}#1@{}}#2\end{tabular}}

\begin{document}

\title{A Fast Point Cloud Ground Segmentation Approach Based on Coarse-To-Fine Markov Random Field}

\author{Weixin Huang, Huawei Liang, Linglong Lin, Zhiling Wang, Shaobo Wang, Biao Yu, Runxin Niu
\thanks{This work has been submitted to the IEEE for possible publication. Copyright may be transferred without notice, after which this version may no longer be accessible.\\
}
\thanks{This work was supported by National Key Research and Development Program of China (Nos. 2016YFD0701401, 2017YFD0700303 and 2018YFD0700602), National Natural Science Foundation of China (Grant No. 91420104), Youth Innovation Promotion Association of the Chinese Academy of Sciences (Grant No. 2017488), Key Supported Project in the Thirteenth Five-year Plan of Hefei Institutes of Physical Science, Chinese Academy of Sciences (Grant No. KP-2019-16), Equipment Pre-research Program (Grant No. 301060603), Natural Science Foundation of Anhui Province (Grant No. 1508085MF133) and Technological Innovation Project for New Energy and Intelligent Networked Automobile Industry of Anhui Province. (Co-corresponding authors: Biao Yu, Runxin Niu)
}
\thanks{The authors are with Hefei Institutes of Physical Science, Chinese Academy of Sciences, Hefei, Anhui, China; University of Science and Technology, Hefei, Anhui, China; Anhui Engineering Laboratory for Intelligent Driving Technology and Application and Innovation Research Institute of Robotics and Intelligent Manufacturing Chinese Academy of Sciences. (e-mail: hwx2018@mail.ustc.edu.cn; hwliang@iim.ac.cn; keith\_lin@rntek.cas.cn;zlwang@hfcas.ac.cn; ba17168@mail.ustc.edu.cn; byu@hfcas.ac.cn; rxniu@iim.ac.cn).}}


\maketitle

\begin{abstract}
Ground segmentation is an important preprocessing task for autonomous vehicles (AVs) with 3D LiDARs. To solve the problem of existing ground segmentation methods being very difficult to balance accuracy and computational complexity, a fast point cloud ground segmentation approach based on a coarse-to-fine Markov random field (MRF) method is proposed. The method uses an improved elevation map for ground coarse segmentation, and then uses spatiotemporal adjacent points to optimize the segmentation results. The processed point cloud is classified into high-confidence obstacle points, ground points, and unknown classification points to initialize an MRF model. The graph cut method is then used to solve the model to achieve fine segmentation. Experiments on datasets showed that our method improves on other algorithms in terms of ground segmentation accuracy and is faster than other graph-based algorithms, which require only a single core of an I7-3770 CPU to process a frame of Velodyne HDL-64E data (in 39.77 ms, on average). Field tests were also conducted to demonstrate the effectiveness of the proposed method.
\end{abstract}


\IEEEpeerreviewmaketitle

\section{Introduction}
\IEEEPARstart{A}{n} accurate and real-time perception of the world is critical for AVs. 3D LiDARs can generate high- resolution 3D point clouds of the environment while remaining unaffected by varying illumination, making them popular and widely used in the perception systems of AVs. In recent years, with the rapid development of key technologies, the price of LiDARs has dropped rapidly, making its application much more valuable.

Given a set of point clouds obtained from a LiDAR sensor, the perception tasks include ground segmentation, 3D object clustering, classification, and tracking. Ground segmentation is a very important step that aims to separate point clouds into obstacle points and ground points to reduce the complexity of subsequent tasks. Mainstream ground segmentation methods for point clouds can be classified into five categories, namely, elevation map methods \cite{r1}, \cite{r2}, ground modeling methods \cite{r3,r4,r5}, time and space adjacent methods \cite{r6,r7}, MRF-based methods \cite{r8,r9,r10,r11,r12,r13}, and deep learning-based methods \cite{r14,r15,r16}.

Elevation map methods project the point cloud into a 2.5D grid map and separate the point cloud into ground points or obstacle points by the height relationship of the points within the grid. Although such methods have been widely used in the past, they assume that the ground in the grid is flat, which makes the segmentation less accurate. Ground modeling methods, though, use mathematical models to model the ground. In the case of large-area modeling, a high-precision ground model can be obtained to further separate the ground points and obstacle points. However, large-area ground modeling is very time consuming and cannot be run in real time. If the range of modeling is reduced, the accuracy is seriously affected. Time and space adjacent methods use the characteristics of the fixed position of lasers in LiDAR systems to quickly complete ground segmentation using information such as the angle between adjacent points. The disadvantage of these methods is that obstacles with a large plane (such as the top of vehicles) will be segmented into the ground, which seriously affects the operation of subsequent tasks. MRF-based methods not only consider the possibility that a single point in a point cloud belongs to a certain classification, but also the continuity of classification between adjacent points, making it a very promising research trend. However, solving MRFs based on iterative algorithms such as belief propagation is slow, and difficult to ensure optimal results. On the other hand, the solution method based on a graph cut requires prior knowledge, which limits its application. Deep-learning-based methods directly segment and classify the point cloud into certain objects through neural networks, and have achieved very good results on public datasets such as SemanticKITTI \cite{r17}. However, this method requires a large amount of training data, and labeling the point clouds is very time consuming, which severely restricts its application in open scenes of AVs.

In view of the above problems, this paper proposes a fast point cloud ground segmentation approach based on coarse-to-fine MRF. The method uses an improved elevation map for ground coarse segmentation, and then uses spatiotemporal adjacent points to optimize the segmentation results. The processed point cloud is classified as high-confidence obstacle points, ground points, and unknown classification points to initialize an MRF model. Then, the graph cut method is used to solve the model to achieve fine segmentation. Compared with the two methods of the elevation map and the time and space adjacent approach, this method uses the MRF model to segment points that are difficult to distinguish, thereby greatly improving the accuracy of the segmentation. Compared with other MRF-based methods, this method uses high-confidence obstacle points and ground points as a priori knowledge, which reduces the computational complexity and helps to directly calculate the global optimal solution without convergence problems.

At present, we have compared this method with other feature extraction algorithms and graph-based algorithms on a public dataset. The results show that our method improves on other algorithms in ground segmentation accuracy and is faster than other graph-based algorithms, which require only a single core of I7-3770 CPU to process a frame of Velodyne HDL-64E data (in 39.77 ms, on average). Field tests were also conducted to demonstrate the effectiveness of the proposed method.

\section{RELATED WORK}
As mentioned before, ground segmentation is an important task for the perception of AVs. According to the types of sensors used, they can be divided into three types: image semantic segmentation, point cloud ground segmentation, and image and point cloud fusion ground segmentation. Under good illumination conditions, the image has rich texture information. An algorithm based on machine learning can label the pixels corresponding to the ground, thus providing reliable road information for the perception system \cite{r18,r19,r20}. However, in an open environment, owing to the varying illumination, the use of images alone cannot guarantee the reliability of ground segmentation. The fusion of 3D point clouds which are less affected by illumination, with images, has been a popular research trend in recent years \cite{r21}. Owing to the richer information, fusion algorithms can often achieve better results than algorithms that use a single type of sensor. However, a large number of computations and the difficulties in real-time matching of point clouds and images limit the further development of fusion algorithms. The amount of data in the point cloud is much smaller than the image, and with the advances of multi-line LiDAR technology, an increasing amount of texture information can be obtained from the point cloud, making LiDAR research key to perception systems. Mainstream ground segmentation methods for point clouds can be classified into five categories: elevation map methods, ground modeling methods, time and space adjacent methods, MRF-based methods, and deep learning-based methods.
\subsection{Elevation Map Methods}
In order to avoid the complexity of full 3D maps, several researchers have considered elevation maps to be an attractive alternative. Ye et al. \cite{r1} proposed an algorithm to acquire an elevation map of a moving vehicle with a tilt laser range scanner. They proposed special filtering algorithms to eliminate measurement errors or noise resulting from the scanner and the motions of the vehicle. To solve the problem of vertical or overhanging objects which cannot be represented correctly, Pfaff et al. \cite{r2} proposed an approach that allows a mobile robot to deal with vertical and overhanging objects in elevation maps, making them more accurate. Although methods based on elevation maps were widely used in the early days of AVs, by neglecting the distribution characteristics of point clouds, an imbalance of points in the grid is caused which affects the accuracy of the algorithm. In addition, this type of algorithm assumes that the ground in the grid is flat, while the ground in the real scene is potentially undulating. To reduce false detection, a sufficiently large threshold needs to be set, which causes  serious problems related to ground over-segmentation.
\subsection{Ground Modeling Methods}
Modeling the ground can improve the accuracy of ground segmentation. Hadsell et al. \cite{r3} proposed an approach that uses online kernel-based learning to estimate a continuous surface over the area of interest while providing upper and lower boundaries on that surface. In this method, visibility information is used to constrain the terrain surface and increase precision, and efficient gradient-based optimization can be achieved quickly. Chen et al. \cite{r4} developed a novel algorithm based on sparse Gaussian process regression (GPR) for segmenting 3D scans of various terrains. The scanned 3D points are first mapped into a 3D grid map, and then the ground is modeled directly using the iterative 2D GPR with sparse covariance functions. However, large-area ground modeling requires many calculations and is difficult to run in real time. In subsequent research, Chen et al. \cite{r5} proposed splitting a large-area ground segmentation problem into many simple GP regression problems with lower complexity, which could then achieve real-time performance while yielding acceptable ground segmentation results. However, the improved algorithm suffered from a problem of classifying stair steps as ground level owing to the smaller fitting area.
\subsection{Time and Space Adjacent Methods}
Ground segmentation through the relationship of adjacent points in space was a common method used in the early days \cite{r6}. However, these algorithms are not considered robust because different seed points usually generate different segments. Considering the fixed position of lasers in LiDAR, Bogoslavskyi et al. \cite{r7} proposed an effective method to remove the ground from the “range image” of a 3d point cloud projection. This method regards the neighboring points in time and space as adjacent points, which can complete the point cloud ground segmentation very quickly and provides a new idea for processing the point cloud. The disadvantage of these methods is that obstacles with a large plane (such as the top of vehicles) are segmented into the ground, which seriously affects the operation of subsequent tasks.
\subsection{MRF-based Methods}
MRF-based point cloud segmentation, which considers not only the possibility that a single point in a point cloud belongs to a certain classification, but also the continuity of classification between adjacent points, is a very promising research trend.
Rusu et al. \cite{r8} proposed an approach for labeling points with different geometric surface primitives. By defining the classes of 3D geometric surfaces and taking advantage of the context information provided by CRF, this method successfully segments and labels 3D points according to their surfaces, even with noisy data. In autonomous driving applications, point cloud data have temporal continuity. Based on this, Lukas et al. \cite{r9} proposed modeling the ground as a spatiotemporal conditional random field (STCRF). Ground elevation parameters are estimated in parallel in each node using an interconnected expectation maximization (EM) algorithm variant. This method successfully segments ground points on various environments with near-real-time performance. However, similar to deep learning methods, such methods have the disadvantage of large computational loads and the need for model pre-training.

Guo et al. \cite{r10} addressed a graph-based approach for 2D road representation of 3D point clouds with respect to road topography. The method also describes the gradient cues of the road geometry to construct an MRF and implements a belief propagation (BP) algorithm to classify the road environment into four categories: reachable, drivable, obstacle, and unknown regions. However, their method uses only gradient values for labeling and sometimes fails to distinguish the ground and roof of a vehicle. With that in mind, Byun et al. \cite{r11} added more point cloud features to the MRF algorithm, making road/obstacle detection more reliable and robust. Zhang et al. \cite{r12} applied the BP method to ground segmentation in a similar manner. While this approach has yielded some research results, the BP method faces three major problems: 1) The calculation of MRF using the BP algorithm is a very expensive computational resource \cite{r22}. \cite{r10,r11,r12} projected the point cloud into a grid map and used the grid as a graph node to reduce the calculation complexity, but the accuracy was correspondingly reduced; 2) in order to distinguish different categories, corresponding labels need to be set. The number of labels is positively correlated with the segmentation results, but the calculation complexity increases dramatically with the number of labels; 3) BP is an iterative method, and it is difficult to ensure that the results converge to a global optimal solution. In summary, there are clear defects in the BP method.

The MRF solution method based on a graph cut \cite{r23} can, however, obtain global optimal solutions. With the help of improved algorithms \cite{r24}, small- and medium-scale graphical model solving can be run in real time. Golovinskiy et al. \cite{r13} presented a min-cut based method for segmenting objects in point clouds. Given an object location, their method builds a k-nearest neighbor graph, assumes a background prior, adds hard foreground (and optionally background) constraints, and finds the min-cut to compute a foreground-background segmentation. This method achieved good segmentation results, but requires prior knowledge of the classification information of a portion of the points, which limits its application in ground segmentation.

\subsection{Deep-Learning-based Methods}
Point cloud segmentation based on deep learning has received considerable attention in recent years. By designing and optimizing the neural network model, the segmentation accuracy has been improved significantly compared to traditional feature-based methods \cite{r14,r15,r16}. However, this method requires a large amount of training data, and labeling the point clouds is very time consuming, which severely restricts its application in open scenes of autonomous vehicles.
Motivated by the analysis of the above methods, we propose a fast point cloud ground segmentation approach based on coarse-to-fine MRF. Inspired by \cite{r2}, we propose a ring-based elevation map method for ground coarse segmentation. Then, a spatiotemporal adjacent points algorithm is proposed to optimize the segmentation results. The processed point cloud is classified as high-confidence obstacle points, ground points, and unknown classification points to initialize an MRF model. Then, the graph cut method is used to solve the model to achieve fine segmentation.

Compared with the two methods of the elevation map and the relationship between adjacent points, this method uses the MRF model to segment the points that are difficult to distinguish, thereby greatly improving the accuracy of the segmentation. Compared with \cite{r10,r11,r12} and other BP methods, this method maps all points in the point cloud to the nodes of the graph, which greatly improves the accuracy. This method uses high-confidence obstacle points and ground points as initialization parameters, which reduces the computational complexity and helps directly calculate the global optimal solution without convergence problems.

The remainder of this paper is organized as follows. Section III presents the details of the fast point cloud ground segmentation approach based on coarse-to-fine MRF. Section IV shows the comparison results of our algorithm with three previous ground segmentation methods \cite{r2,r7,r12} on the public dataset in \cite{r17,r25}. The field test results are also given. Finally, conclusions and an outline of future work are presented in Section V.

\section{COARSE TO FINE GROUND SEGMENTATION}
\subsection{Framework of Algorithm}
\begin{figure}[htbp]
\centering
\includegraphics[width=3.5in]{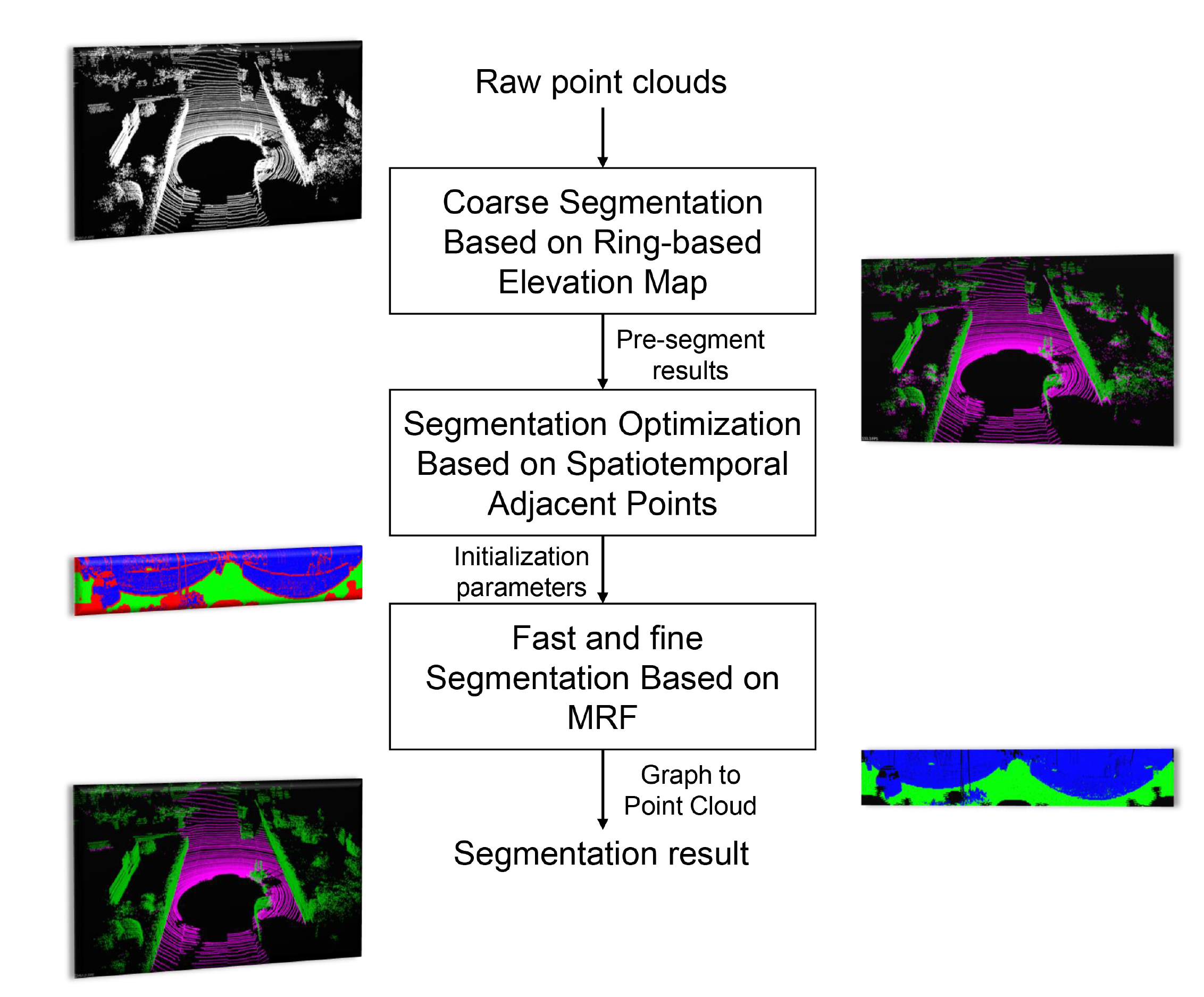}
\caption{Algorithm framework.}
\label{fig_1}
\end{figure}

Our algorithm framework is shown in Fig. \ref{fig_1}. They are coarse segmentation based on a ring-based elevation map, segmentation optimization based on spatiotemporal adjacent points, and fast and fine segmentation based on MRF. The details of each part are given in detail below.

\subsection{Coarse Segmentation Based on the Ring-based Elevation Map}
Elevation map methods, such as \cite{r1}, \cite{r2} ignore the distribution characteristics of the point cloud, resulting in an unbalanced distribution of point clouds in the grid, which ultimately affects the accuracy of the algorithm.

\begin{figure}[htbp]
\centering
\includegraphics[width=3in]{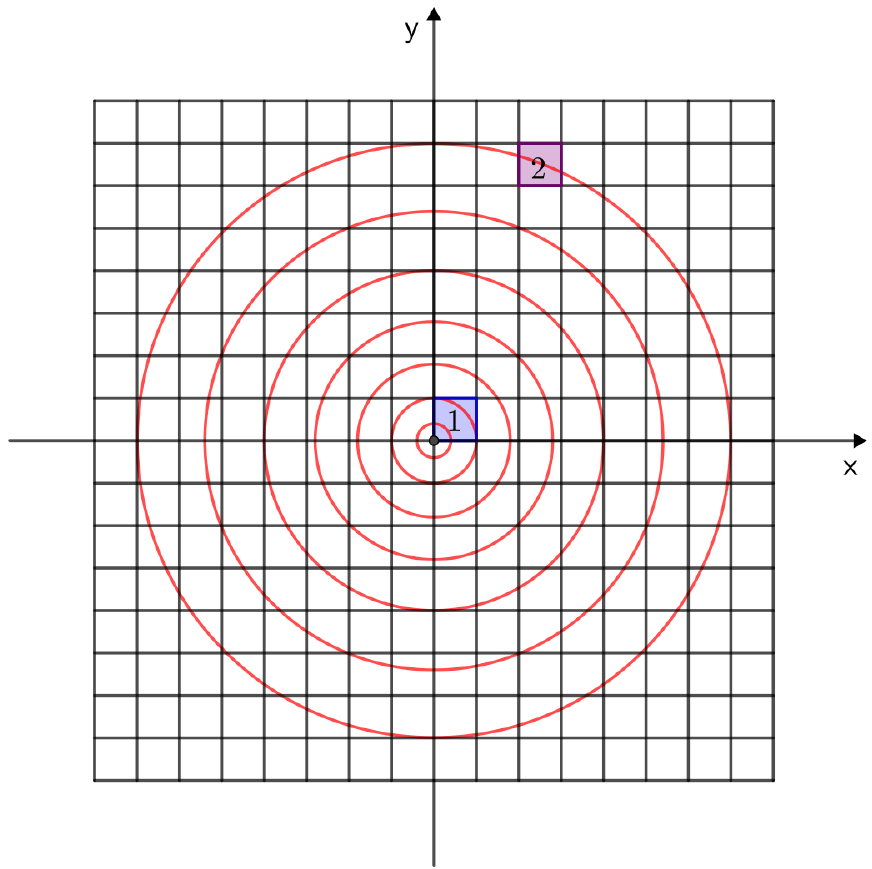}
\caption{Dividing a grid at a fixed x/y distance in a bird's eye view of a point cloud to build an elevation map.}
\label{fig_2}
\end{figure}

The point cloud is a collection of points emitted by multiple lasers. As shown in Fig. \ref{fig_2}, the points corresponding to different lasers are arranged in different rings in the bird's eye view, and these points become sparse as the distance from the origin of the coordinates increases. Dividing the grid with a fixed x/y distance, the number of points in the grid is as shown in Equation \ref{e1}.

\begin{equation}
N_{t}=\sum_{i=0}^{L}\left\{\begin{array}{cl}
P_{\mathrm{N}}(i), & \text { if } y_{2}<\sqrt{R_{i}^{2}-x_{1}^{2}}<y_{1},\vspace{2ex} \\  &\quad\quad  y_{2}<\sqrt{R_{i}^{2}-x_{2}^{2}}<y_{1} \vspace{2ex} \\ 
0, & \text { otherwise }
\end{array}\right.
\label{e1}
\end{equation}

\begin{equation}
P_{\mathrm{N}}(i)=P_{\mathrm{t}} * \frac{\left|\tan ^{-1} \frac{x_{1}}{\sqrt{R_{i}^{2}-x_{1}^{2}}}-\tan ^{-1} \frac{x_{2}}{\sqrt{R_{i}^{2}-x_{2}^{2}}}\right|}{2\pi}
\label{e2}
\end{equation}

where $R_i$ is the radius of the circle formed by the points corresponding to the laser $i$ in the bird's eye view, $x_1$, $x_2$ are the horizontal coordinate values of the left and right boundaries of the grid, respectively, $y_1$ and $y_2$ are the vertical coordinate values of the upper and lower boundaries of the grid, respectively, and $P_t$ is the total number of points obtained by a single laser scan.

Taking grid 1 and grid 2 shown in Fig. \ref{fig_2} as an example, $N_{t1}=\frac{P_t}{2}$,$N_{t2}=0.017P_t$, there is a nearly 30 times difference in the number of points contained in the two grids.

\begin{figure}[htbp]
\centering
\includegraphics[width=3.5in]{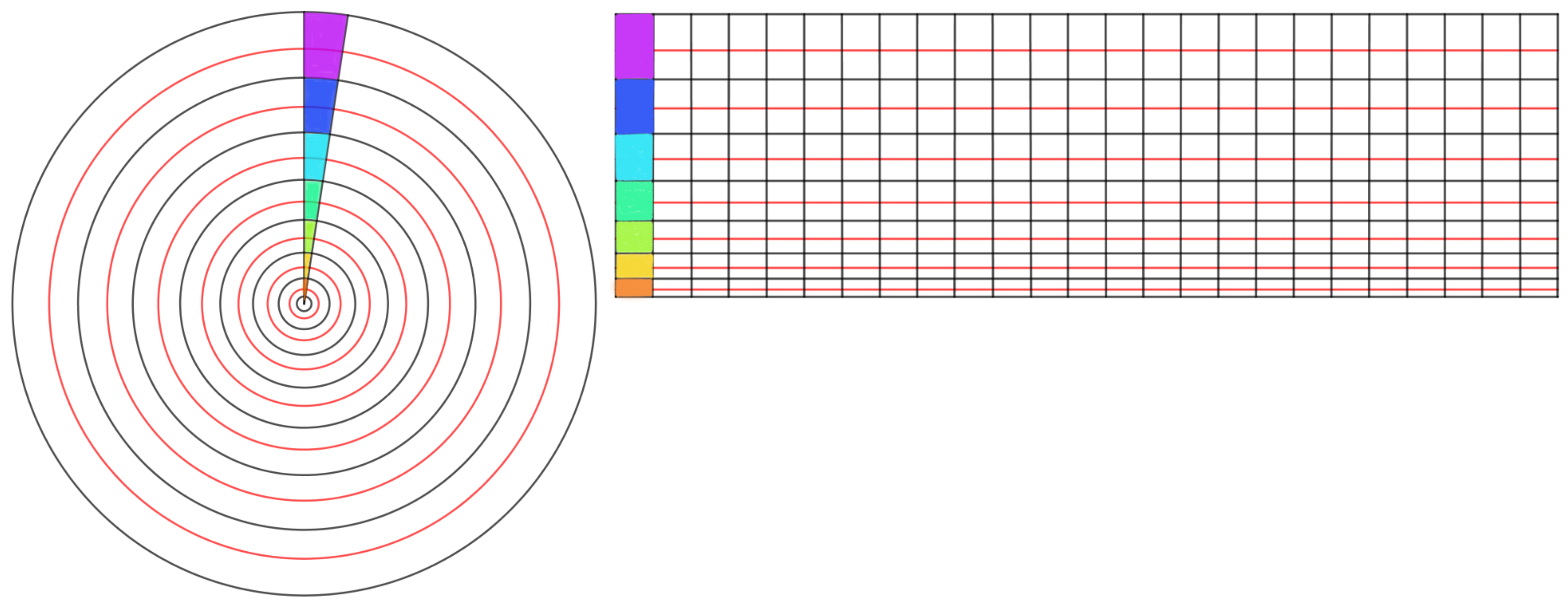}
\caption{Ring-based grid partitioning, where the red line represents the point cloud distribution location and the black line represents the grid's longitudinal boundary line.}
\label{fig_3}
\end{figure}

When there are obstacles around the vehicle, the distribution of the point cloud changes. If the ground point cannot be obtained in the grid, the obstacle points will be incorrectly segmented into ground points. That is, in the traditional elevation map ground segmentation algorithm, the algorithm's error probability increases with the distance of the obstacle. To solve this problem, our study proposes a ring-based elevation map method for ground segmentation, as shown in Fig. \ref{fig_3}.

In the bird's eye view, the point cloud unfolds according to the angle between the point and the vertical axis. The horizontal axis of the unfolded plot is the angle between the point and the vertical axis of the top view, and the vertical axis is the distance between the point and the coordinate origin in the aerial view. In the unfolded plot, the grid is divided by taking the fixed distance as the horizontal boundary and the average distance (obtained by calibration) between the corresponding points of adjacent lasers and the origin in the aerial view as the vertical boundary. This method of grid division makes the possibility of the ground points falling into different grids closer, and based on this, using the elevation map method for ground segmentation is more accurate.

\subsection{Segmentation Optimization Based on Spatiotemporal Adjacent Points}
The ring-based elevation map method for obstacle point extraction assumes that the ground in the grid is flat, while the ground in a real scene is undulating. To reduce false detection, a sufficiently large threshold must be set, which causes a serious ground over-segmentation problem. To mitigate this problem, a segmentation optimization algorithm based on spatiotemporal adjacent points is proposed.

\begin{figure}
\centering
\subfigure[]{\label{fig_4_a}
\includegraphics[width=0.45\linewidth]{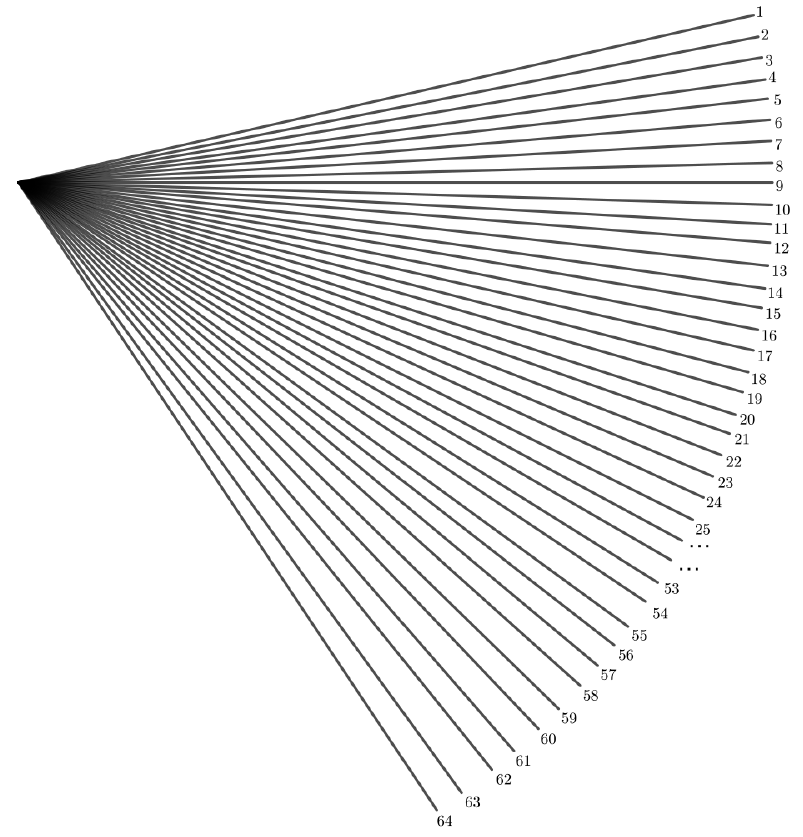}}
\hspace{0.01\linewidth}
\subfigure[]{\label{fig_4_b}
\includegraphics[width=0.45\linewidth]{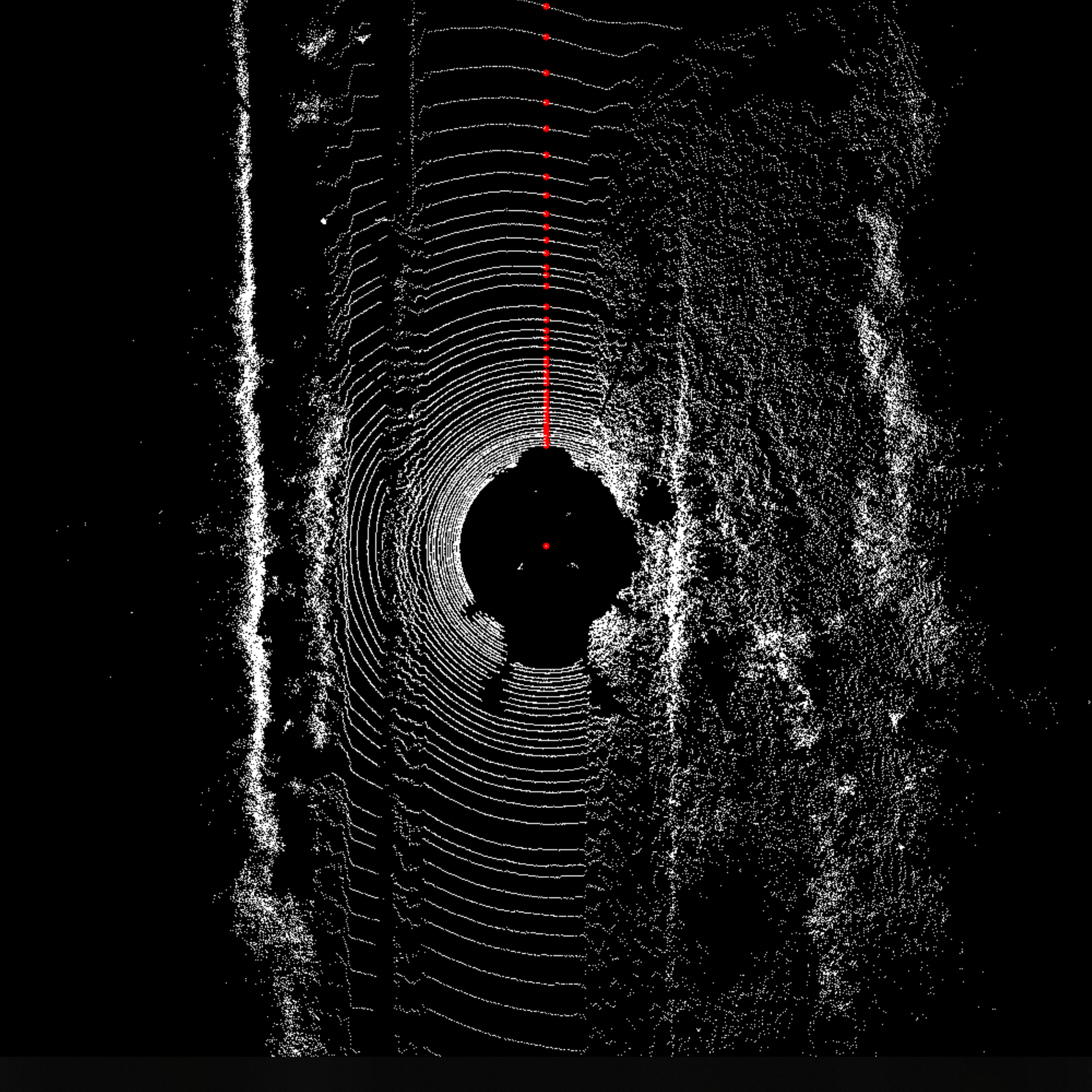}}
\caption{(a) Side view of points at the same horizontal angle. (b) Bird's-eye view of points (red points) at the same horizontal angle.}
\label{fig_4}
\end{figure}

The main LiDAR used for autonomous driving navigation is usually mounted on the roof of the vehicle to scan the environment around the vehicle. Inside the LiDAR, the lasers are arranged in a straight or staggered manner, so that the relative horizontal angle between the lasers is fixed. After calibration, the points of all lasers in the LiDAR at the same horizontal angle can be obtained. The side view of these points is shown in Fig. \ref{fig_4}a, and the bird's-eye view is shown in Fig. \ref{fig_4}b.

\begin{figure}
\centering
\subfigure[]{\label{fig_5_a}
\includegraphics[width=0.45\linewidth]{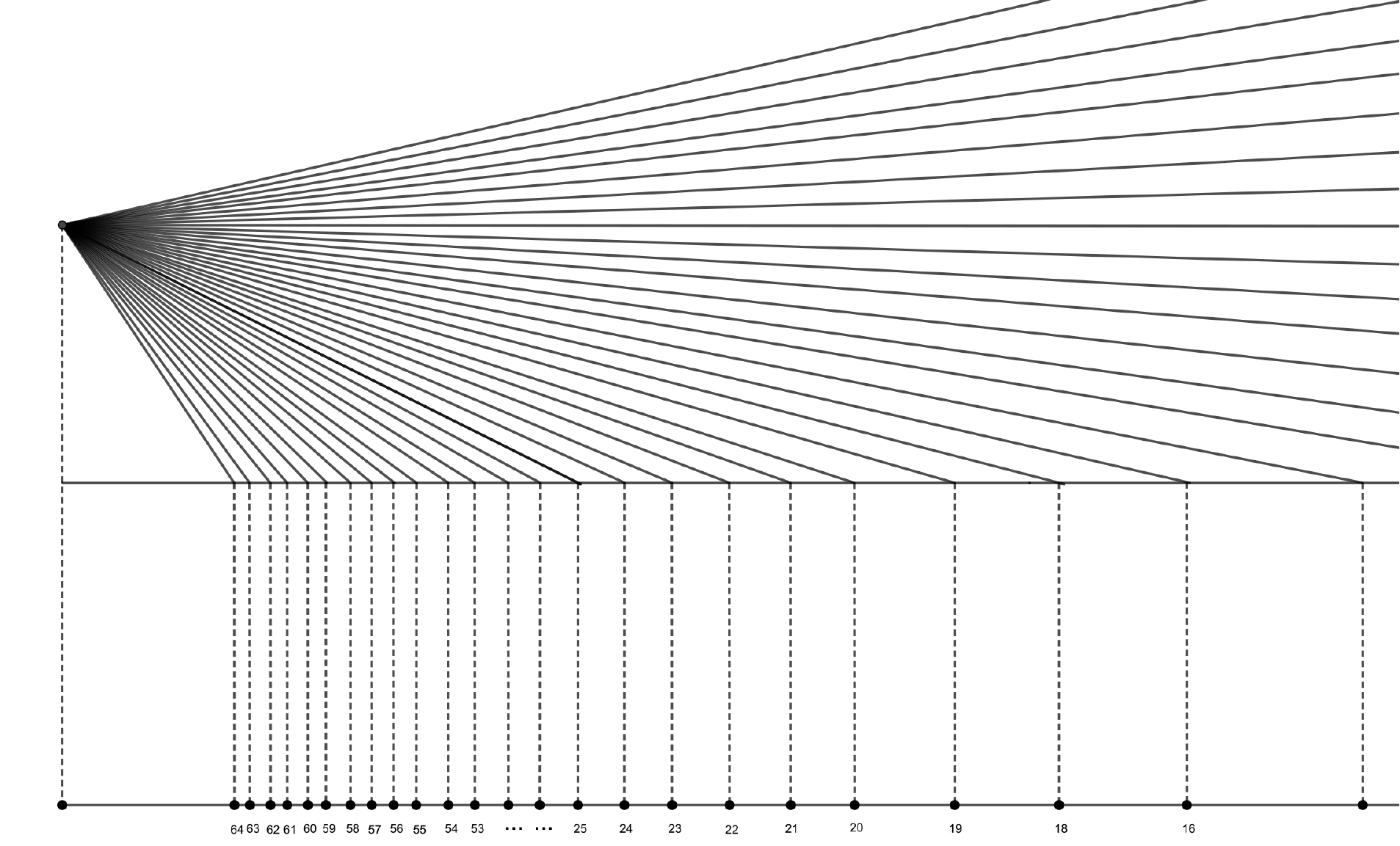}}
\hspace{0.01\linewidth}
\subfigure[]{\label{fig_5_b}
\includegraphics[width=0.45\linewidth]{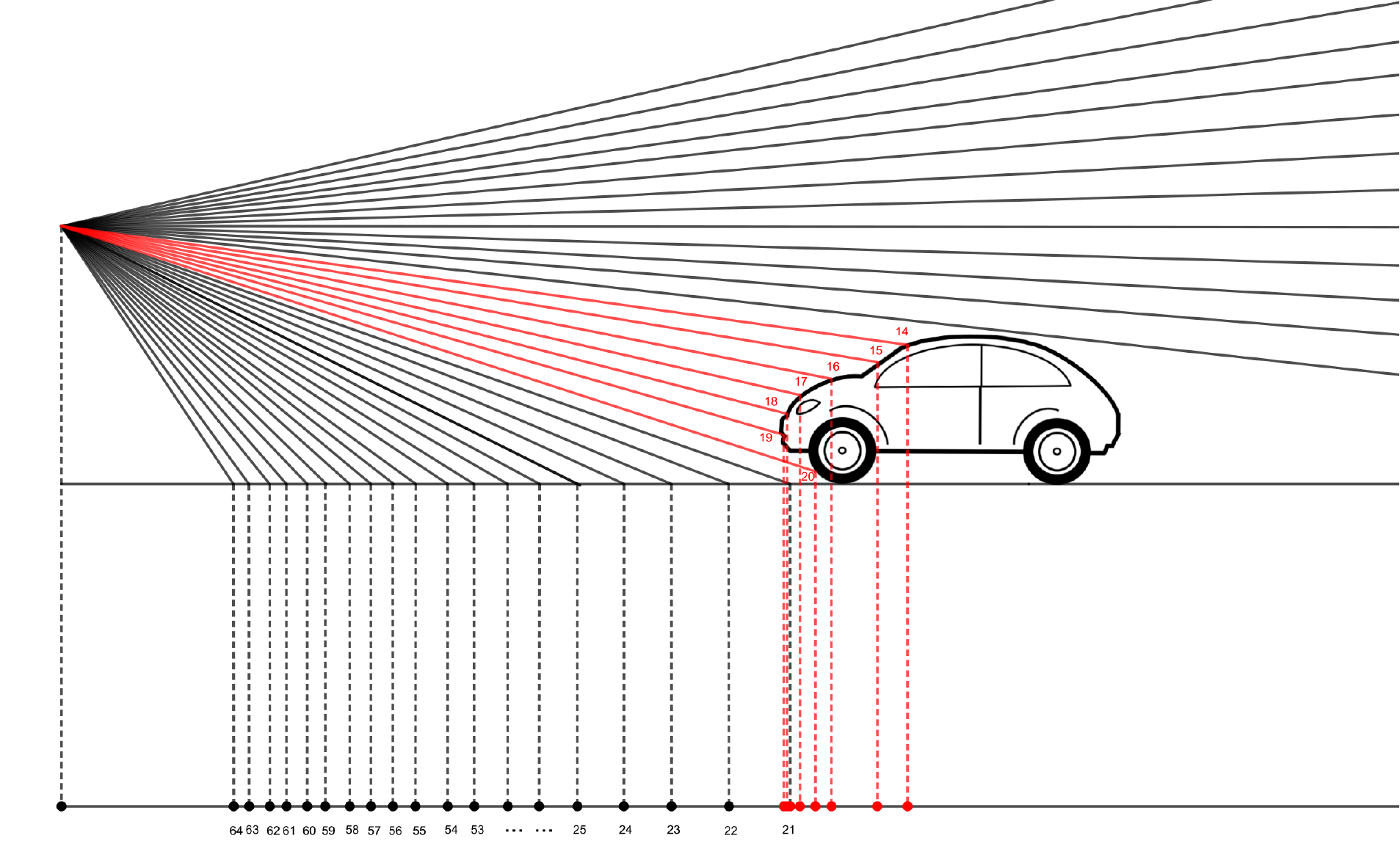}}
\caption{The arrangement of points in two cases: (a) There are no obstacles. (b) There are obstacles.}
\label{fig_5}
\end{figure}

When there are no obstacles in the scanning direction, the horizontal distance between the point and the origin increases with the vertical angle of its corresponding laser (Fig. \ref{fig_5}a). When obstacles exist, the arrangement and distance relationship of these points changes accordingly (Fig. \ref{fig_5}b).

\begin{figure}[htbp]
\centering
\includegraphics[width=2.5in]{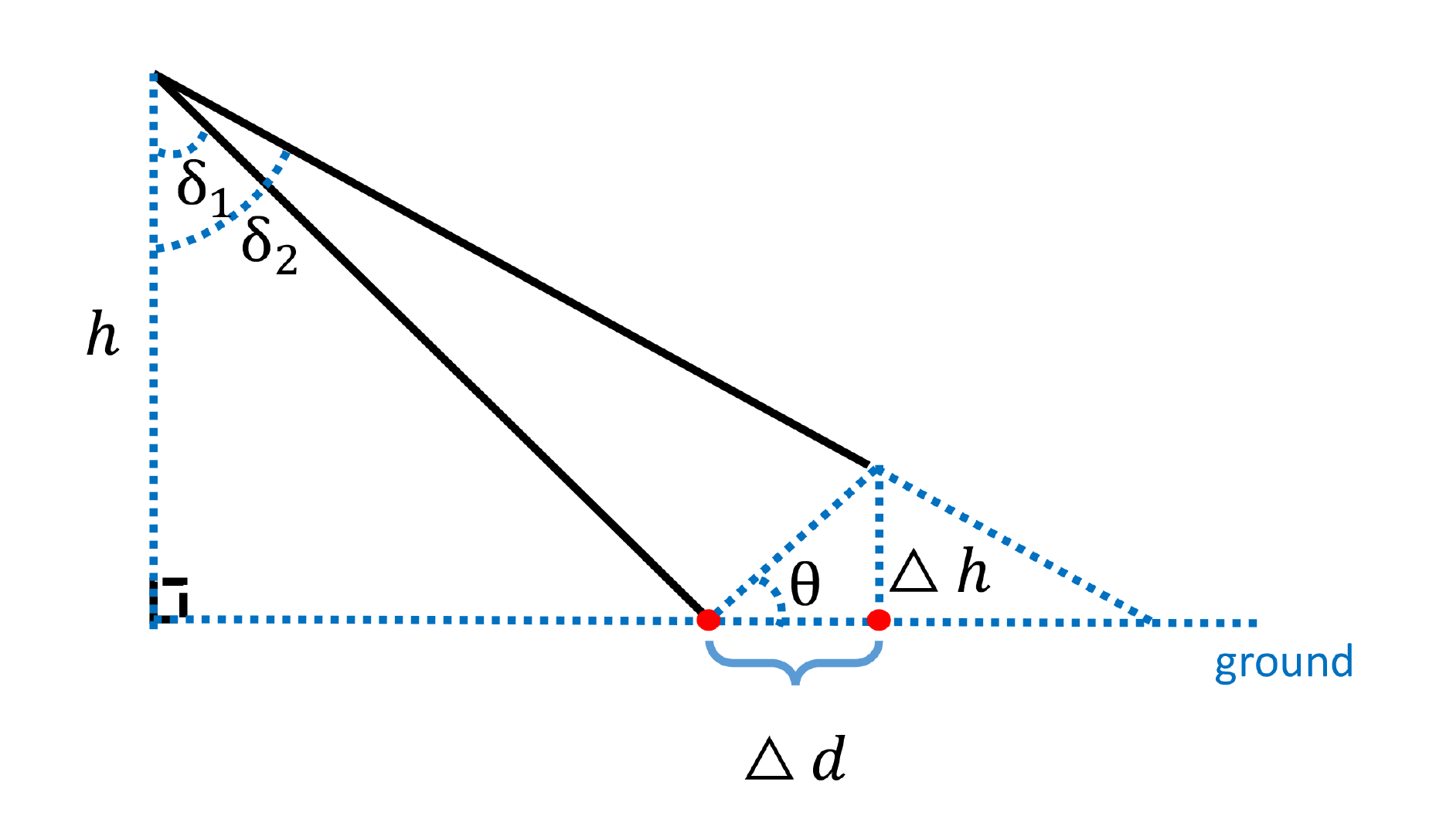}
\caption{Relationship between the corresponding points of different lasers in the same horizontal direction and the height fluctuation.}
\label{fig_6}
\end{figure}

As shown in Fig. \ref{fig_6}, the height fluctuation in the horizontal direction can be calculated according to the distance relationship between the corresponding points of different lasers in the same horizontal direction in the bird's-eye view.

The height difference $\triangle h$ between the corresponding points of the two lasers can be obtained by Equation \ref{e3}, and the elevation angle between them can be obtained by Equation \ref{e4}.

\begin{equation}
\Delta h=\frac{h\left(\tan \delta_{2}-\tan \delta_{1}\right)-\Delta d}{\tan \delta_{2}}
\label{e3}
\end{equation}

\begin{equation}
\theta=\arctan \frac{\Delta h}{\Delta d}
\label{e4}
\end{equation}

where h is the LiDAR mounting height, $\delta_1$ and $\delta_2$ are the angles between the laser and the ground normal.

Because the maximum height difference and elevation angle allowed on the ground can be set artificially, if the maximum value of $\theta$ is set to $K$, then,

\begin{equation}
\Delta d \geq \frac{h\left(\tan \delta_{2}-\tan \delta_{1}\right)}{\operatorname{tanKtan} \delta_{2}+1}
\label{e5}
\end{equation}

\begin{figure}[htbp]
\centering
\includegraphics[width=2.5in]{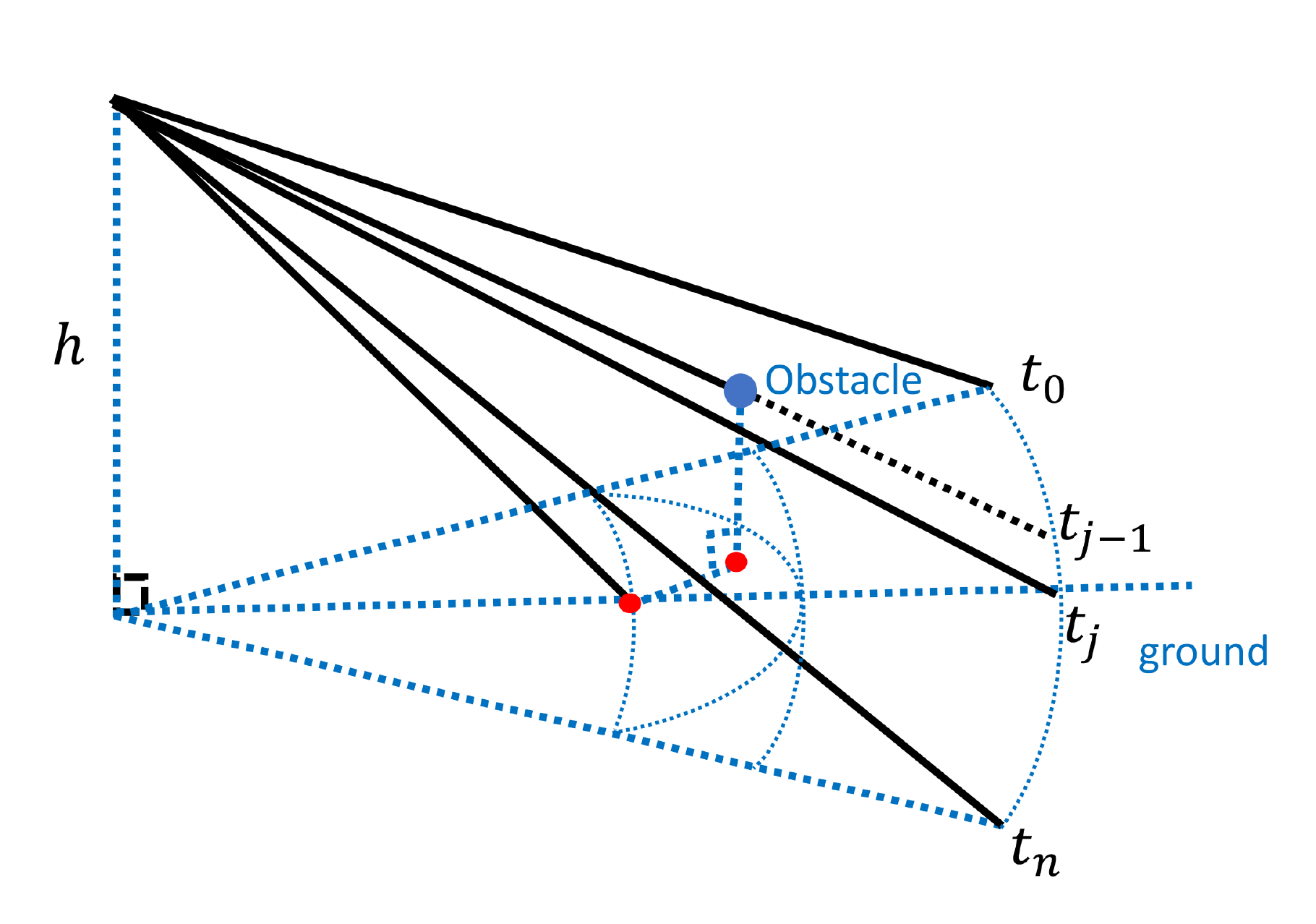}
\caption{Relationship between the corresponding points of different lasers in the close horizontal direction and the height fluctuation.}
\label{fig_7}
\end{figure}

In other words, if there are other points in the same horizontal direction within the $\triangle d$ horizontal distance of one point, there are obstacles in these points. Furthermore, because the update of the point cloud data is continuous, the conclusion can be generalized to points of a close horizontal angle (Fig. \ref{fig_7}).

This algorithm converts trigonometric function calculations into distance calculations, which is very efficient. Most obstacle points can be quickly extracted through the specific algorithm shown in Algorithm 1 to mitigate the problem of ground over-segmentation based on the ring-based elevation map ground segmentation method.

\begin{algorithm}[htbp]
    \caption{Obstacle cluster and closing}
    \begin{algorithmic}[1]
    \REQUIRE
      $Raw\ Point\ cloud$\\
    \ENSURE
      $Marked\ Point\ cloud$\\
	
	\STATE $colselzoo \gets 3$
	\STATE $LINE \gets 64$
	\STATE $LPQUAN \gets 2000$
		\FOR{$l=0$ to $LINE-2$}
			\FOR{$p=colselzoo$ to $LPQUAN-colselzoo$}
				\FOR{$sp=p-colselzoo$ to $sp+colselzoo$}
					\STATE $mptdistance \gets PDIS(pCloud[l][p], pCloud[l + 2][sp])$
					\IF {$mptdistance < LineMaxDistance[l]$}
						\STATE $ pCloud[l + 2][sp].marktype \gets OBS$
					\ENDIF
				\ENDFOR
			\ENDFOR
		\ENDFOR
		\RETURN{$pCloud$}

    \end{algorithmic}
\end{algorithm}

\subsection{Fast and Fine Segmentation Based on MRF}

The two methods proposed above can quickly complete the ground segmentation of point clouds. However, because segmentation only considers local features, the problem of over-segmentation remains. To further solve this problem, this paper proposes fast and fine segmentation based on MRF.

Unlike \cite{r8,r9,r10} and other methods that use grids as graph nodes to reduce the computational complexity, in order to achieve more accurate segmentation results, this paper proposes a graph construction method that projects all points in the point cloud to graph nodes.

\begin{figure}[htbp]
\centering
\includegraphics[width=3.5in]{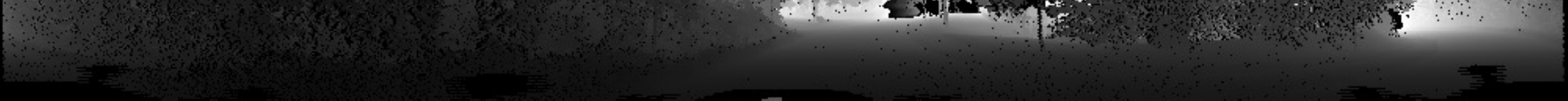}
\caption{Range image. The number of rows in the image is defined by the number of lasers in the LiDAR. The number of columns is given by the laser's range reading of 360° per rotation. The Velodyne HDL-64E, for example, has a range map width of about 1800 and a height of 64.}
\label{fig_8}
\end{figure}

First, the point cloud data is directly converted to a range image based on the laser vertical and horizontal angle information provided by the LiDAR device manufacturer, as shown in Fig. \ref{fig_8}. Then, we construct the graph $G=(P,N)$ with the pixels of the range image as the nodes of the graph and the 8-connected relations of the pixels as the connection of the graph. Finally, we transform the ground segmentation problem into a two-class labeling problem in the graph, and the quality of the classification is constrained by the energy function,

\begin{equation}
E(A)=\lambda \cdot R(A)+B(A)
\label{e6}
\end{equation}
where,
\begin{equation}
R(A)=\sum_{p \in \mathcal{P}} R_{p}\left(A_{p}\right)
\label{e7}
\end{equation}

\begin{equation}
B(A)=\sum_{(p, q) \in N} B_{(p, q)} \cdot \delta\left(A_{p}, A_{q}\right)
\label{e8}
\end{equation}

and,

\begin{equation}
\delta\left(A_{p}, A_{q}\right)=\left\{\begin{array}{l}
1 \text { if } A_{p} \neq A_{q} \\
0 \text { otherwise }
\end{array}\right.
\label{e9}
\end{equation}

The regional term $R\left(A\right)$ comprises the individual penalties for assigning graph node p to “background” and “obstacle”. The term $B\left(A\right)$ comprises the “boundary” properties of segmentation $A$. The coefficient $B_{\left\{p,q\right\}}\geq\ 0$ should be interpreted as the penalty for a discontinuity between $p$ and $q$. The coefficient $\lambda\geq0$ in Equation \ref{e6} specifies the relative importance of the region properties term $R(A)$ versus the boundary properties term $B\left(A\right)$.

\begin{figure}[htbp]
\centering
\includegraphics[width=3.5in]{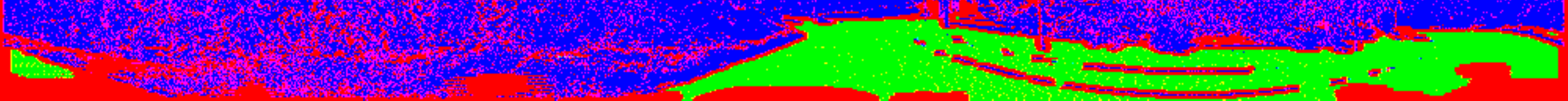}
\caption{Initialized MRF model, in which the green part is the “background” node, the blue part is the “obstacle” node, and the red part is the node to be segmented.}
\label{fig_9}
\end{figure}

The graph-based method has a black-box characteristic and requires a large degree of calculation. In order to reduce the impact of these two problems and obtain the global optimal segmentation results, this paper proposes a method to initialize the MRF model using the feature algorithm and solve it with graph cuts. Using the graph cut method requires knowing the classification of some nodes in advance \cite{r26}. Based on this, we consider the obstacle points obtained by the feature extraction algorithm described above as high-confidence obstacle points to initialize the “object” node. This operation ensures that the high-confidence obstacle points are not erroneously segmented into ground points during the graph cut process and provides a lot of obstacle space information for the MRF model. The ground points obtained by the feature extraction algorithm include obstacle points that have been incorrectly segmented, so they cannot be used directly to initialize the “background” node. Observation of the range image reveals that most of the ground points are connected into blocks. Suppose that in a region centered on a point, if the proportion of pre-segmented ground points is greater than $k$, the point is regarded as a ground point with high degree of confidence. Using high-confidence ground points to initialize their corresponding nodes as “background” nodes, results in the graph structure shown in Fig. \ref{fig_9}.

Fig. \ref{fig_9} shows that most nodes have been initialized as “objects” or “backgrounds,” thus greatly simplifying the calculation of the model. Even if the number of nodes in the graph is much larger than the method using grids as graph nodes, the method can still run in real time. After initializing the MRF model, we define the regional term $R\left(A\right)$ and the boundary term $B\left(A\right)$ of the energy function of MRF model based on the distribution characteristics of the point cloud. Referring to the application of the histogram in image processing tasks \cite{r27}, the height of the point cloud is enlarged proportionally and then rounded to $g(A_p)$,

\begin{equation}
g\left(A_{p}\right)=\left[\left(h\left(A_{p}\right)+h_{l}\right) * k\right]
\label{e10}
\end{equation}

where $h_l$ is the value of the lowest point on the ground in the point cloud, and $k$ is the scale factor.
Performing histogram statistics on the $g(A_p)$ values of the nodes initialized to “objects” and “backgrounds” to obtain the probability density functions $D_{obj} (g(A_p))$ and $D_{bkg} (g(A_p))$, then we define the regional term $R(A)$ according to Equation \ref{e11}.

\begin{equation}
R\left(A\right)=\sum_{p\in"bkg"}{R_p\left(A_p\right)}+\sum_{p\in"obj"}{R_p\left(A_p\right)}\ \ \ \ \ 
\label{e11}
\end{equation}
where,
\begin{equation}
\sum_{p\in"bkg"}{R_p\left(A_p\right)}=-\sum_{p\in"obj"}{ln(D_{obj}(g(A_p))}
\label{e12}
\end{equation}

\begin{equation}
\sum_{p\in"obj"}{R_p\left(A_p\right)}=-\sum_{p\in"bkg"}{{ln(D}_{obj}(g(A_p))}
\label{e13}
\end{equation}

The boundary term $B\left(A\right)$ is related to the gradient between adjacent points, and we define it using Equation \ref{e15}.

\begin{equation}
B_{(p, q)}=\exp \left(-\sigma \frac{|h(p)-h(q)|^{2}}{d(p, q)}\right)
\label{e14}
\end{equation}

where $h\left(p\right)$and $h\left(q\right)$ are the heights of adjacent points, $d\left(p,q\right)$ is the distance between two points in the horizontal direction, and $\sigma$ is the adjustment coefficient.

\begin{equation}
B\left(A\right)=\sum_{\left\{p,q\right\}\in\mathcal{N}}{\exp{\left(-\sigma\frac{\left|h\left(p\right)-h\left(q\right)\right|^2}{d\left(p,q\right)}\right)}\cdot\delta\left(A_p,A_q\right)}
\label{e15}
\end{equation}

\begin{figure}[htbp]
\centering
\subfigure[]{\label{fig_10_a}
\includegraphics[width=0.45\linewidth]{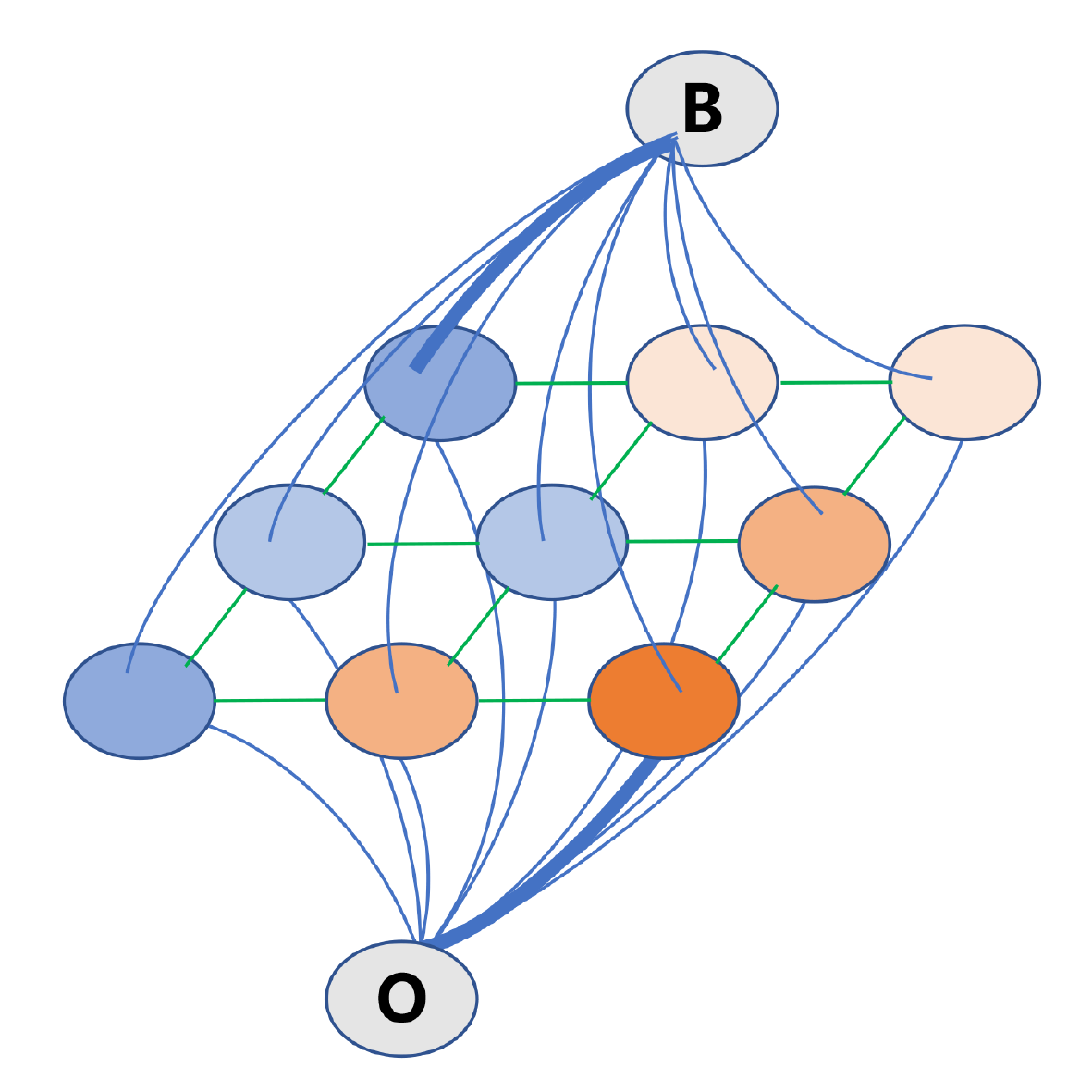}}
\hspace{0.01\linewidth}
\subfigure[]{\label{fig_10_b}
\includegraphics[width=0.45\linewidth]{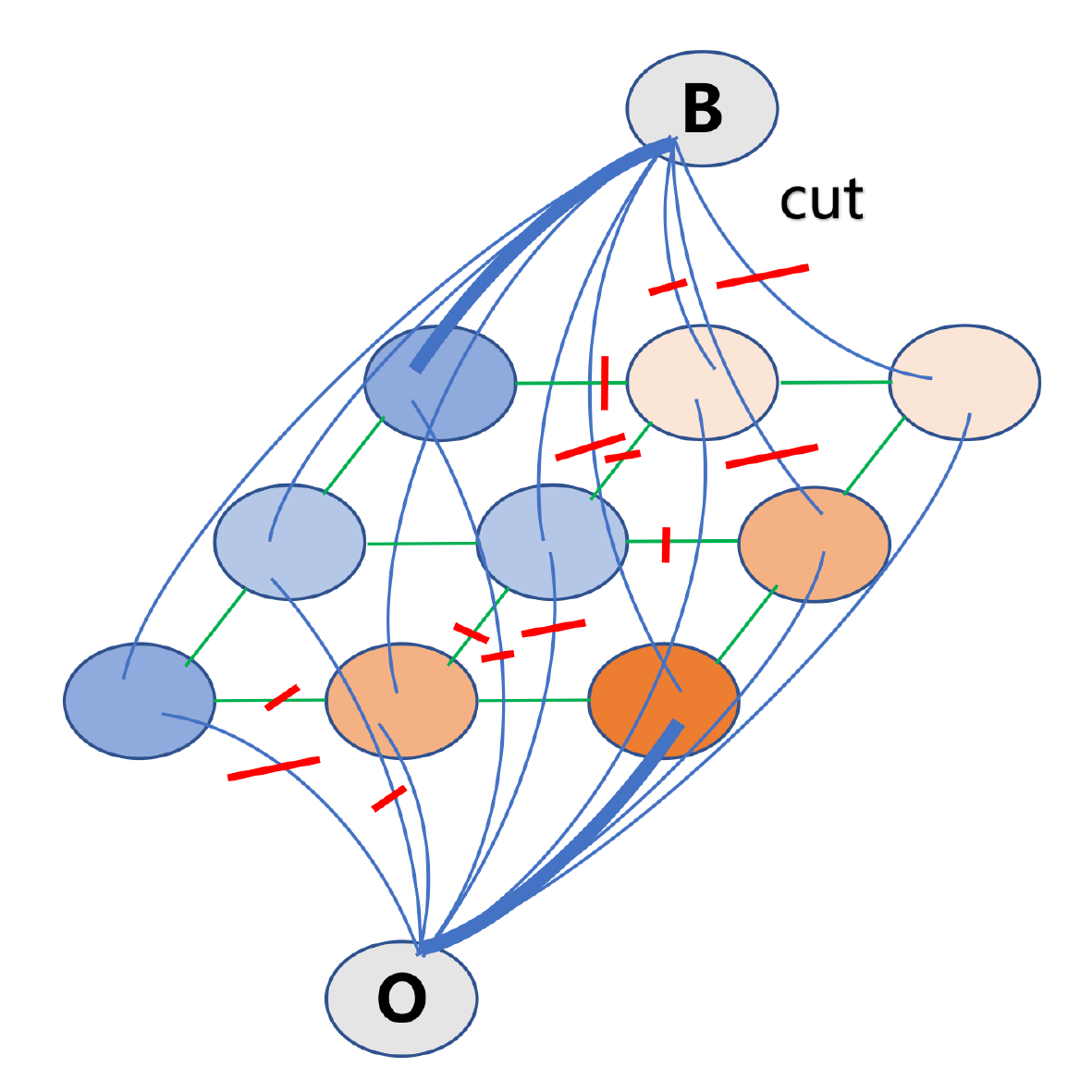}}
\hspace{0.01\linewidth}
\subfigure[]{\label{fig_10_c}
\includegraphics[width=0.45\linewidth]{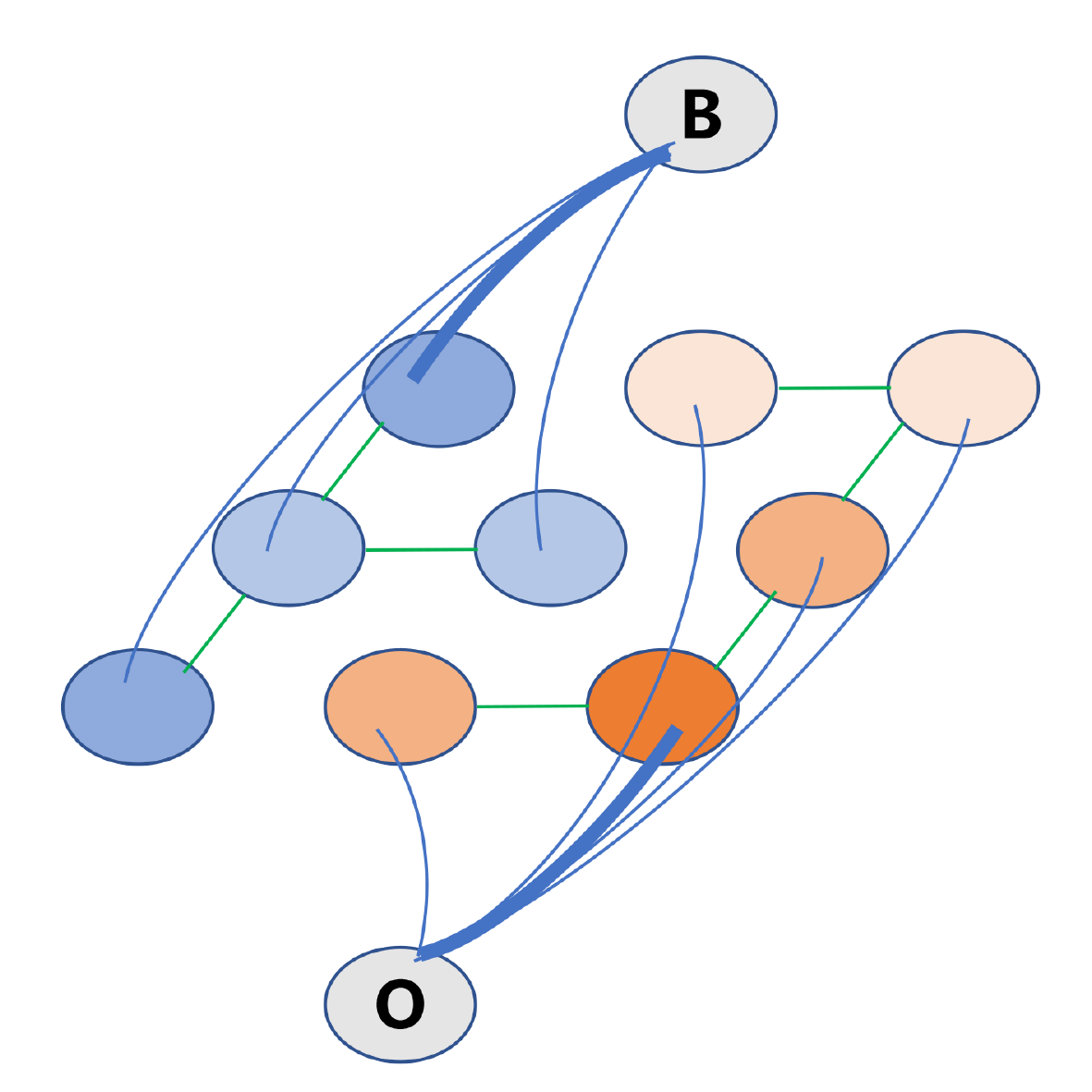}}
\caption{(a) MRF model (b) Min-cut. (c) MRF model after removing the min-cut edge.}
\label{fig_10}
\end{figure}

By this point, we have completed all the required definitions and obtained the MRF model shown in Fig. \ref{fig_10}a. Through the fast algorithm proposed in \cite{r24}, the min-cut corresponding to the minimum energy $E\left(A_{min}\right)$ can be obtained, as shown in Fig. \ref{fig_10}b.

The MRF model after removing the min-cut edge is shown in Fig. \ref{fig_10}c. The nodes connected to the “object” endpoint are obstacle nodes, and the nodes connected to the “background” endpoint are ground nodes. The precise ground segmentation results can be obtained from the mapping relationship between the graph nodes and the point clouds.

\section{EXPERIMENTAL AND DISCUSSION}
Using the ground segmentation methods proposed in \cite{r2,r7,r12} as a comparison, the proposed algorithm was tested based on the following four aspects: 1) the improvement of the ground segmentation results by each module in the method, 2) the overall performance of the method, 3) the stability of the method in different environments, and 4) the real-time processing ability of the method. The method running speed was affected by the choice of programing language and computer hardware, so we implemented all methods in C++ and tested the methods on the same computer (an Intel i7-3770 CPU and 16 GB RAM). We then ran the algorithm on a self-developed autopilot platform to evaluate how well it actually worked.

\subsection{Experimental Datasets}
We selected the SemanticKITTI \cite{r17} dataset for experiments with common scenarios and the Koblenz \cite{r25} dataset for experiments with complex scenarios.

\begin{figure}
\centering
\subfigure[]{\label{fig_11_a}
\includegraphics[width=0.98\linewidth]{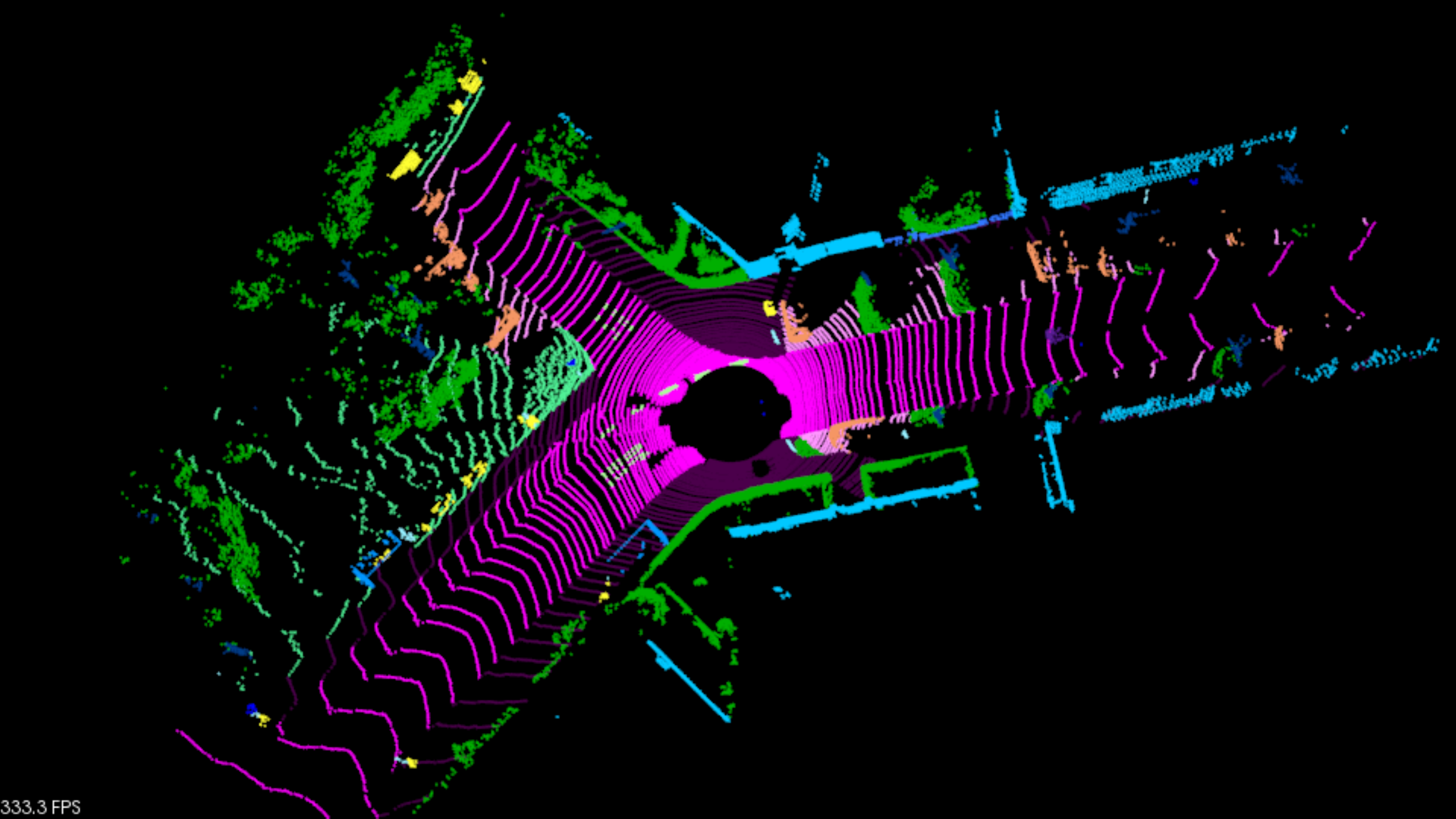}}
\hspace{0.01\linewidth}
\subfigure[]{\label{fig_11_b}
\includegraphics[width=0.98\linewidth]{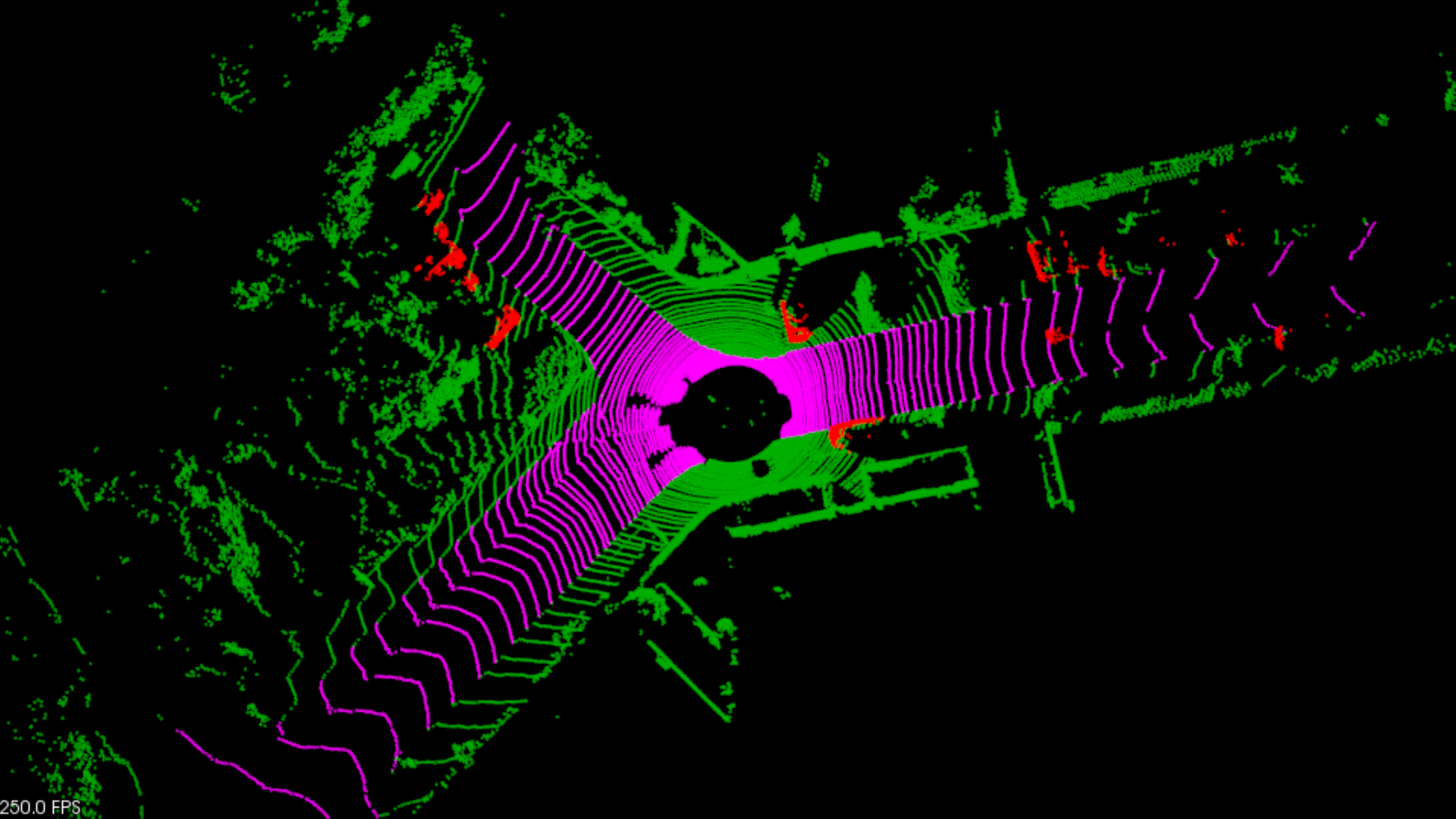}}
\caption{(a) The SemanticKITTI dataset provides 25 classification labels, and different color points represent different classifications. (b) The labels of the SemanticKITTI dataset are merged into three classifications: “ground”, “ordinary obstacle” and “key obstacle.”}
\label{fig_11}
\end{figure}

The SemanticKITTI dataset contains 11 sequences of labeled data, including 23,201 full 3D scans (point clouds) of city traffic, residential areas, and highway scenes, as shown in Table I. It was collected using Velodyne HDL-64E, providing labeling of 25 classes, as shown in Fig. \ref{fig_11}a. As the detection of pedestrians, bicycles and motor vehicles is closely related to the safety of autonomous driving, we merge them into “key obstacles” and the other classifications into “ground” and “ordinary obstacle”, the processed data is shown in Fig. \ref{fig_11}b.

\begin{table*}[htbp]
\renewcommand{\arraystretch}{1.3}
\caption{THE 11 SEQUENCES WITH LABELS IN THE SEMANTIC KITTI DATASET}
\centering
\begin{tabular}{ccccccccccccc}
\hline
sequences&00&01&02&03&04&05&06&07&08&09&10\\
\hline
 scenarios& \tabincell{c}{residential\\area}&\tabincell{c}{highway\\scene}&\tabincell{c}{city\\traffic}&\tabincell{c}{residential\\area}&\tabincell{c}{city\\traffic}&\tabincell{c}{residential\\area}&\tabincell{c}{city\\traffic}&\tabincell{c}{residential\\area}&\tabincell{c}{residential\\area}&\tabincell{c}{city\\traffic}&\tabincell{c}{city\\traffic}\\
scans &4540&1100&4660&800&270&2760&1100&1100&4070&1590&1200\\							
 \hline
\end{tabular}
\end{table*}

\begin{figure}
\centering
\subfigure[]{\label{fig_12_a}
\includegraphics[width=0.98\linewidth]{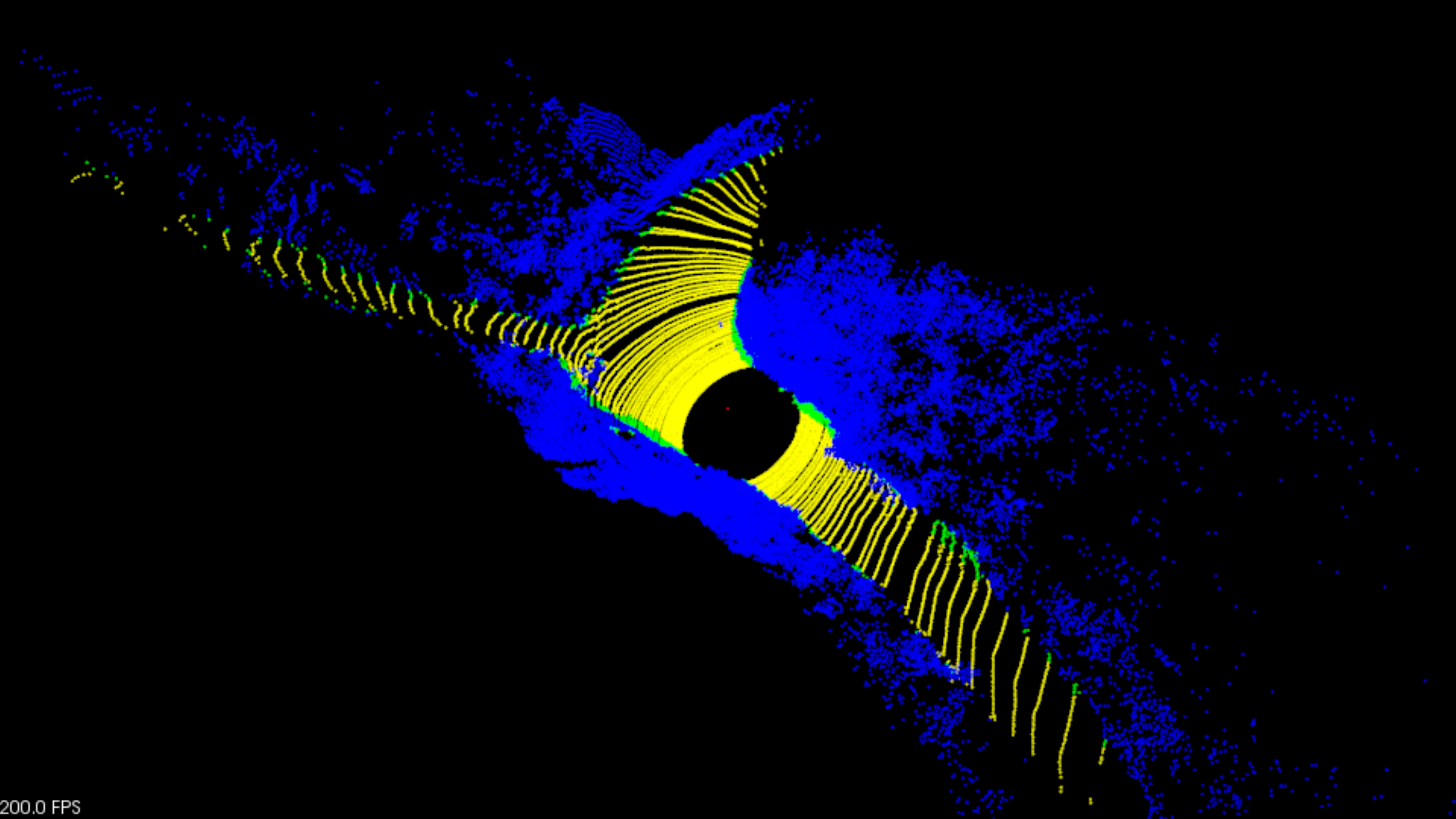}}
\hspace{0.01\linewidth}
\subfigure[]{\label{fig_12_b}
\includegraphics[width=0.98\linewidth]{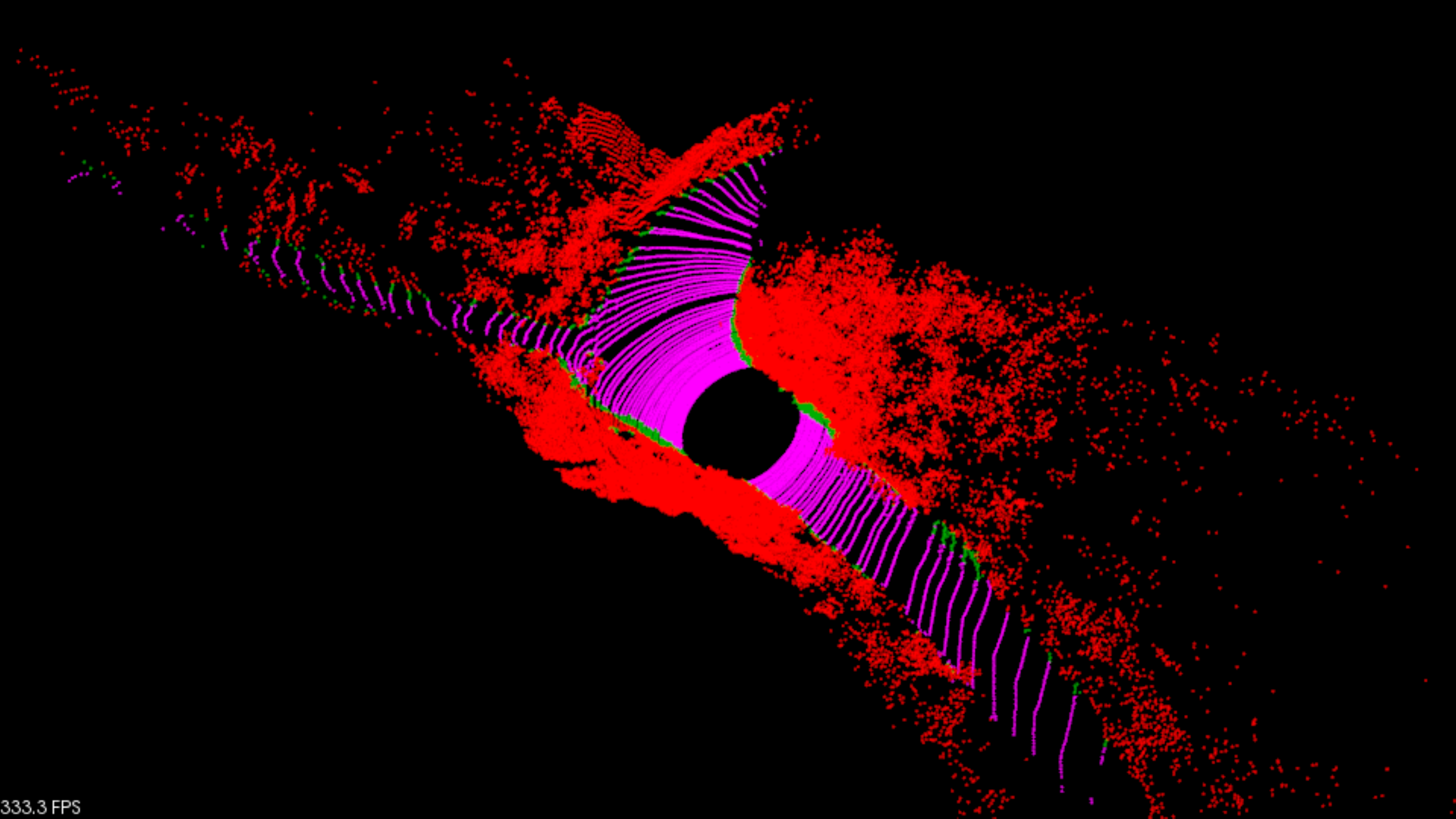}}
\caption{(a) The Koblenz dataset provides 4 classification labels, and different color points represent different classifications. (b) The labels on the Koblenz dataset were converted into three classifications: “ground”, “ordinary obstacle” and “key obstacle.”}
\label{fig_12}
\end{figure}

The Koblenz dataset contains three subsets: KoblenzForestMap, KoblenzCampusMap, and KoblenzFarmMap. The KoblenzForestMap was recorded along a sloped forest road in the municipal forest of Koblenz, which is characterized by many trees and the fact that the forest terrain surface is completely occluded. The KoblenzCampusMap consists of over 64 million 3D points from the university campus and is characterized by buildings, asphalt, and road segments, some meadows, and small vegetated areas. The KoblenzFarmMap was created on a farm road with many fields, meadows, and vegetation. This scenario is the most complex due to rainy weather, puddles, translucent vegetation, and vibrations from the dirt road. A summary of the three scenarios is shown in Table II. This dataset, also collected by the Velodyne HDL-64E, provides labeling of four classes: “Street”, “Obstacle”, “Rough” and “Unknown”, as shown in Fig. \ref{fig_12}a. In order to unify with the organized SemanticKITTI dataset, we converted "Street” to “ground”, “Rough” to “ordinary obstacles” and “Obstacle” to “key obstacles.” The processed data is shown in Fig. \ref{fig_12}b.

\begin{table*}[htbp]
\renewcommand{\arraystretch}{1.3}
\caption{THE KOBLENZ DATASET}
\centering
\begin{tabular}{cccc}
\hline
Koblenz dataset&KoblenzForestMap&KoblenzCampusMap&KoblenzFarmMap\\
\hline
scenarios&forest&campus&farm\\
scans&263&644&426\\						
 \hline
\end{tabular}
\end{table*}

Unlike the comparison algorithm, our algorithm requires point and laser correspondence information in order to convert point cloud data into range images. However, the SemanticKITTI dataset does not retain this information, so we backtracked it. A small number of errors occurred due to the lack of information during the backtracking process, which caused the performance of our method to degrade in the dataset. The Koblenz dataset, however, does not have this problem. In addition, for the convenience of calculation, the corresponding points of different lasers in the LiDAR system were aligned in the horizontal direction in advance.

\subsection{Experimental Evaluation}
As shown in Equation \ref{e16}, we use intersection-over-union (IoU) \cite{r28} to evaluate the segmentation results of the ground points.

\begin{equation}
I o u_{g}=\frac{T P_{g}}{T P_{g}+F P_{g}+F N_{g}}
\label{e16}
\end{equation}

where $T P_{g}$, $F P_{g}$, and $F N_{g}$ correspond to the number of true positive, false positive, and false negative predictions for class “ground.”
Because the ground segmentation method cannot guarantee complete accuracy, some obstacle points may be incorrectly segmented into ground points, leading to unstable point cloud clustering and even traffic accidents in extreme cases. Since pedestrians, bicycles, and motor vehicles in ordinary scenarios, rigid obstacles such as forests, farms, and campuses in complex scenarios, are closely related to the safety of autonomous driving, we consider them as “key obstacles” and use the recall rate shown in Equation \ref{e17} to evaluate their segmentation results.

\begin{equation}
\text {Recall}_{o}=\frac{T P_{o}}{T P_{o}+F N_{o}}
\label{e17}
\end{equation}

where $T P_{o}$ and $F N_{o}$ correspond to the number of true positive and false negative predictions for class “key obstacles.”

\begin{table*}[htbp]
\renewcommand{\arraystretch}{1.3}
\caption{AVERAGE PERFORMANCE COMPARISON TABLE}
\centering
\begin{tabular}{cccc}
\hline
Method&$Iou_g$&$Recall_o$&time(ms)\\
\hline
Pfaff \cite{r2}&37&64.11&8.94\\
Ring-based elevation map method&39.83&81.16&\textbf{7.68}\\	
Ring-based elevation map method + Spatiotemporal adjacent points method&43.73&85.92&13.02\\	
Ring-based elevation map method + Spatiotemporal adjacent points method + MRF method&\textbf{48.58}&\textbf{93.71}&39.04\\						
 \hline
\end{tabular}
\end{table*}

\begin{figure}[htbp]
\centering
\includegraphics[width=3.5in]{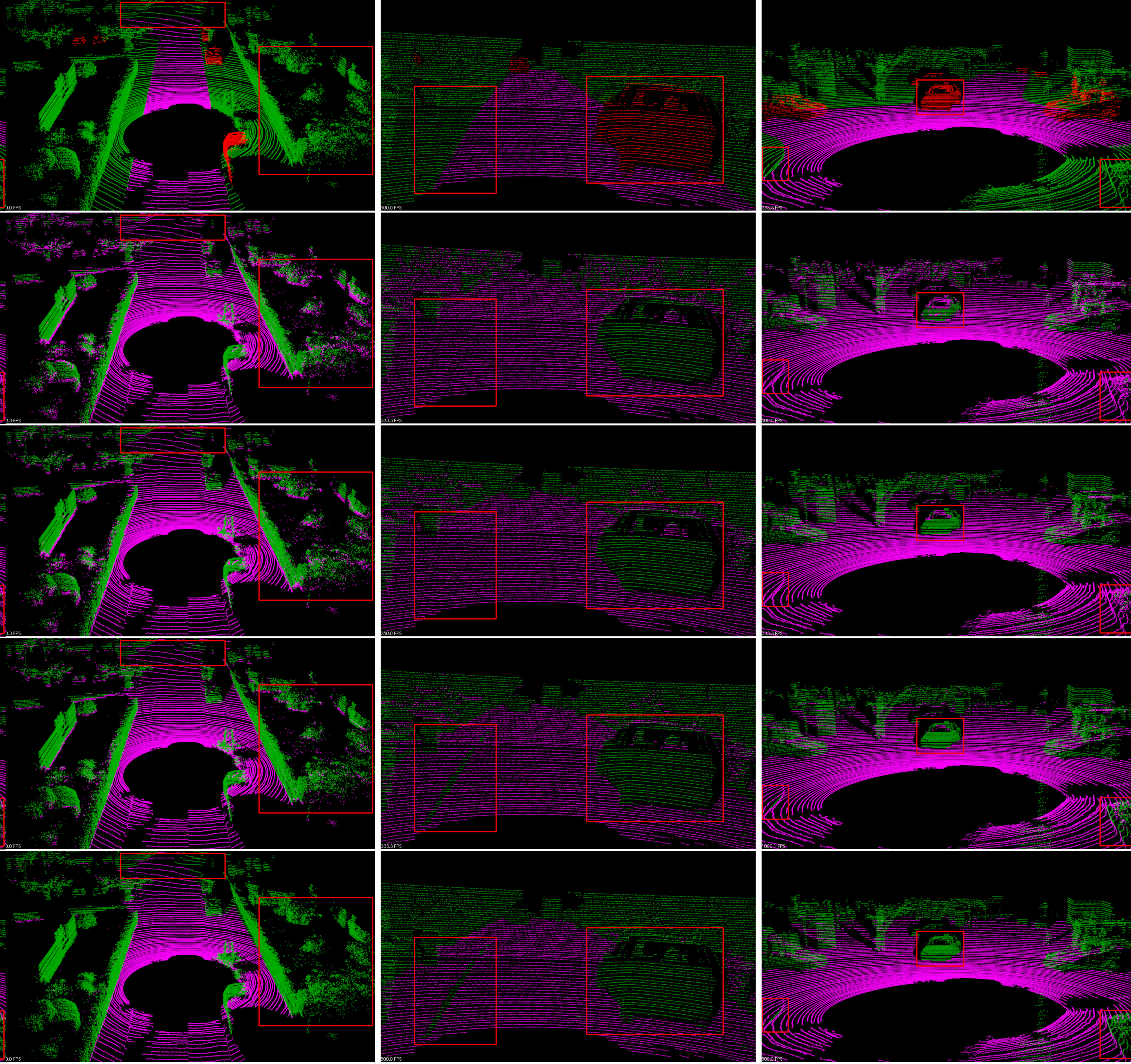}
\caption{In three different scenes, Some qualitative comparison results obtained by the method in \cite{r2} (the second line), ring-based elevation map method (the third line), ring-based elevation map method + spatiotemporal adjacent points method (the fourth line), and ring-based elevation map method + spatiotemporal adjacent points method + MRF method (the fifth line). The first line is the benchmark for point cloud classification, where red, green, and pink points represent “key obstacles,” “ordinary obstacles,” and “ground,” respectively.}
\label{fig_13}
\end{figure}

Table III illustrates the average performance of each module of our proposed algorithm in the test scenarios. In order to verify the performance of the ring-based elevation map method, we used the method proposed by \cite{r2} as a comparison. As shown in Table III, the ring-based elevation map method makes the possibility of ground points falling into different grids closer, and consequently more accurate. Notably, the recall rate of “key obstacles” increased significantly (17.05 \%), which improves the safety of autonomous driving. The difference between the two methods can be seen in the second and third lines of Fig. \ref{fig_13}. The spatiotemporal adjacent points method is more sensitive to gradient and can quickly search for some obstacle points that have been incorrectly segmented into the ground, thereby further improving the accuracy of ground point segmentation. More importantly, it can detect some low road edges, making the distinction between inside and outside the road more accurately, as shown in the fourth line of Fig. \ref{fig_13}. 

As graph-based algorithms have global characteristics, the two evaluation indexes are further improved with the addition of the MRF method. By comparing the fourth and fifth lines of Fig. \ref{fig_13}, it is found that most of the over-segmentation caused by the small calculation range was corrected. The above experiments show that each module of the proposed method significantly improves the segmentation results. In order to obtain the overall performance of the proposed method in this study, we compared it with the method proposed in \cite{r7}, [12] across all scenarios in the two datasets.

The experimental results for 11 scenarios in the SemanticKITTI dataset are shown in Tables IV, V, and VI.

\begin{table*}[htbp]
\renewcommand{\arraystretch}{1.3}
\caption{STATISTICAL RESULTS OF $Iou_{g}$ ON SEMANTICKITTI DATASET (\%)}
\centering
\begin{tabular}{ccccccccccccc}
\hline
Method&Seq00&Seq01&Seq02&Seq03&Seq04&Seq05&Seq06&Seq07&Seq08&Seq09&Seq10&mean\\
\hline
Zhang \cite{r12}&34.02&58.16&34.71&33.53&55.19&30.8&30.34&36.81&31.46&35.73&32.04&37.53\\
Bogoslavskyi \cite{r7}&23.96&51.63&26.85&29.29&45.58&22.84&28&25.35&25.04&28.17&21.86&29.87
\\	
Our&\textbf{43.67}&\textbf{68.54}&\textbf{43.38}&\textbf{43.89}&\textbf{63.31}&\textbf{38.84}&\textbf{34.03}&\textbf{47.47}&\textbf{41.56}&\textbf{46.2}&\textbf{47.26}&\textbf{47.10}\\					
 \hline
\end{tabular}
\end{table*}

\begin{table*}[htbp]
\renewcommand{\arraystretch}{1.3}
\caption{STATISTICAL RESULTS OF $Recall_{o}$ ON SEMANTICKITTI DATASET (\%)}
\centering
\begin{tabular}{ccccccccccccc}
\hline
Method&Seq00&Seq01&Seq02&Seq03&Seq04&Seq05&Seq06&Seq07&Seq08&Seq09&Seq10&mean\\
\hline
Zhang \cite{r12}&72.09&61.21&74.84&72.32&80.32&76.06&78.63&74.71&73.46&76.18&71.66&73.77\\
Bogoslavskyi \cite{r7}&58.76&54.36&53.72&50.98&53.81&58.28&55.86&58.16&55.13&56.06&60.02&55.92\\	
Our&\textbf{96.35}&\textbf{93.3}&\textbf{94.91}&\textbf{96.2}&\textbf{97.92}&\textbf{96.55}&\textbf{97.23}&\textbf{96.64}&\textbf{96.53}&\textbf{93.25}&\textbf{94.09}&\textbf{95.72}\\		
 \hline
\end{tabular}
\end{table*}

\begin{table*}[htbp]
\renewcommand{\arraystretch}{1.3}
\caption{STATISTICAL RESULTS OF RUNTIMES ON SEMANTICKITTI DATASET (MS)}
\centering
\begin{tabular}{ccccccccccccc}
\hline
Method&Seq00&Seq01&Seq02&Seq03&Seq04&Seq05&Seq06&Seq07&Seq08&Seq09&Seq10&mean\\
\hline
Zhang \cite{r12}&1477.63&1482.06&1475.77&1476.36&1479.85&1477.14&1487.17&1473.15&1476.5&1473.06&1469.47&1477\\
Bogoslavskyi \cite{r7}&\textbf{15.94}&\textbf{14.88}&\textbf{15.79}&\textbf{16.96}&\textbf{16.32}&\textbf{16.16}&\textbf{15.72}&\textbf{15.10}&\textbf{15.41}&\textbf{15.53}&\textbf{15.62}&\textbf{15.8}\\	
Our&39.59&37.77&39.21&42.09&39.94&39.83&40.82&38.98&39.69&40.67&38.89&39.8\\
 \hline
\end{tabular}
\end{table*}

As shown in Tables IV and V, the method proposed in this paper has the best performance in all test scenarios. The recall rate indicator of “key obstacles,” which is closely related to the safety of autonomous driving, performed particularly well, with the average performance of all scenarios exceeding 93\%. The method in \cite{r7} has a serious over-segmentation problem owing to the limitation of the calculation range, so its detection accuracy was the worst. Compared with the former, the graph-based method in [12] showed better performance in continuous space. However, in the presence of occlusion, there may be a lack of spatial continuity points around the obstacle for reference, resulting in some points that were obviously higher than the ground being incorrectly segmented into the ground. Our method obtained a lot of obstacle point information from the segmentation results of the feature algorithm, to overcome this problem. In terms of runtime, Table VI shows that the method \cite{r7} had the fastest processing speed. The method proposed in this paper does a lot of optimization in the construction and solution of MRF model, so it runs much faster than the graph-based method in [12] and can run in real time.

The Koblenz dataset contains a large amount of irregular ground and is a major challenge for ground segmentation. The experimental results for the three scenarios in the Koblenz dataset are shown in Tables VII, VIII, and IX.

\begin{table}[htbp]
\renewcommand{\arraystretch}{1.3}
\caption{STATISTICAL RESULTS OF $Iou_g$ ON KOBLENZ DATASET (\%)}
\centering
\begin{tabular}{ccccc}
\hline
Method&ForestMap&CampusMap&FarmMap&mean\\
\hline
Zhang \cite{r12}&36.87&62.96&24.93&41.59\\
Bogoslavskyi \cite{r7}&24.37&44.81&20.40&29.86\\	
Our&\textbf{58.89}&\textbf{73.65}&\textbf{29.51}&\textbf{54.02}\\
 \hline
\end{tabular}
\end{table}

\begin{table}[htbp]
\renewcommand{\arraystretch}{1.3}
\caption{STATISTICAL RESULTS OF $Recall_o$ ON KOBLENZ DATASET (\%)}
\centering
\begin{tabular}{ccccc}
\hline
Method&ForestMap&CampusMap&FarmMap&mean\\
\hline
Zhang \cite{r12}&64.01&74.62&57.50&65.38\\
Bogoslavskyi \cite{r7}&47.15&72.26&55.83&58.42\\	
Our&\textbf{89.38}&\textbf{90.81}&\textbf{78.76}&\textbf{86.32}\\
 \hline
\end{tabular}
\end{table}

\begin{table}[htbp]
\renewcommand{\arraystretch}{1.3}
\caption{STATISTICAL RESULTS OF RUNTIMES ON KOBLENZ DATASET (MS)}
\centering
\begin{tabular}{ccccc}
\hline
Method&ForestMap&CampusMap&FarmMap&mean\\
\hline
Zhang \cite{r12}&1507.88&1501.65&1522.74&1510.76\\
Bogoslavskyi \cite{r7}&\textbf{15.08}&\textbf{15.76}&\textbf{15.13}&\textbf{15.32}\\
Our&34.78&35.84&38.43&36.35\\	
 \hline
\end{tabular}
\end{table}

\begin{figure}[htbp]
\centering
\includegraphics[width=3.5in]{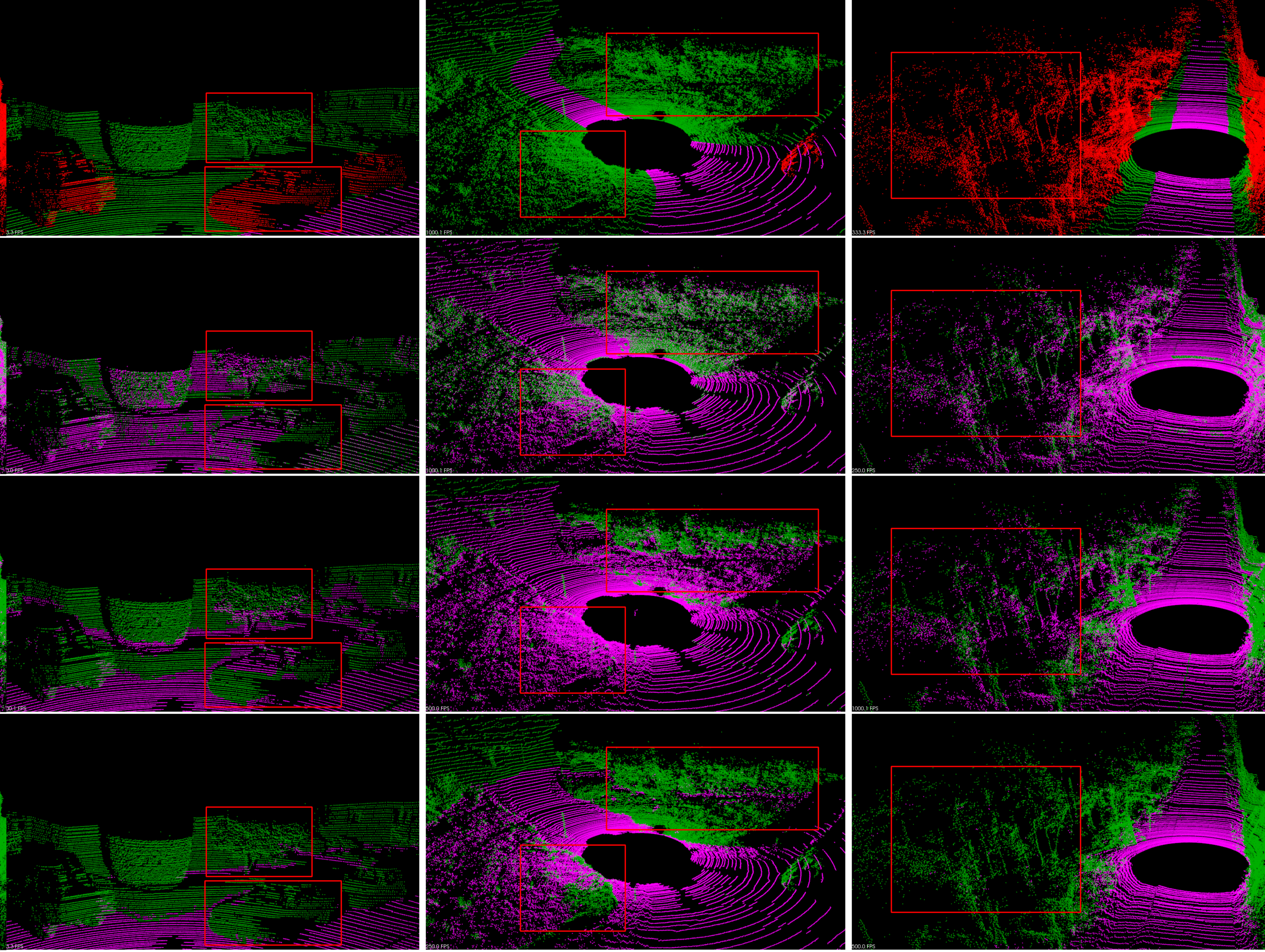}
\caption{In three different scenes, some qualitative comparison results obtained by the method in \cite{r2} (the second line), the method in [12] (the third line), and the method in this paper (the fourth line) in three different scenes. The first line is the benchmark for point cloud classification, where red, green, and pink points represent “key obstacles,” “ordinary obstacles,” and “ground,” respectively.}
\label{fig_14}
\end{figure}

As can be seen from Tables VII, VIII, and IX, the algorithm proposed in this paper also achieves the best performance in all complex scenarios. In terms of runtime, the performances of the three algorithms are consistent with their performance in ordinary scenarios. Some of the experimental results of the three methods are shown in Fig. \ref{fig_14}.

\begin{figure}[htbp]
\centering
\includegraphics[width=3.5in]{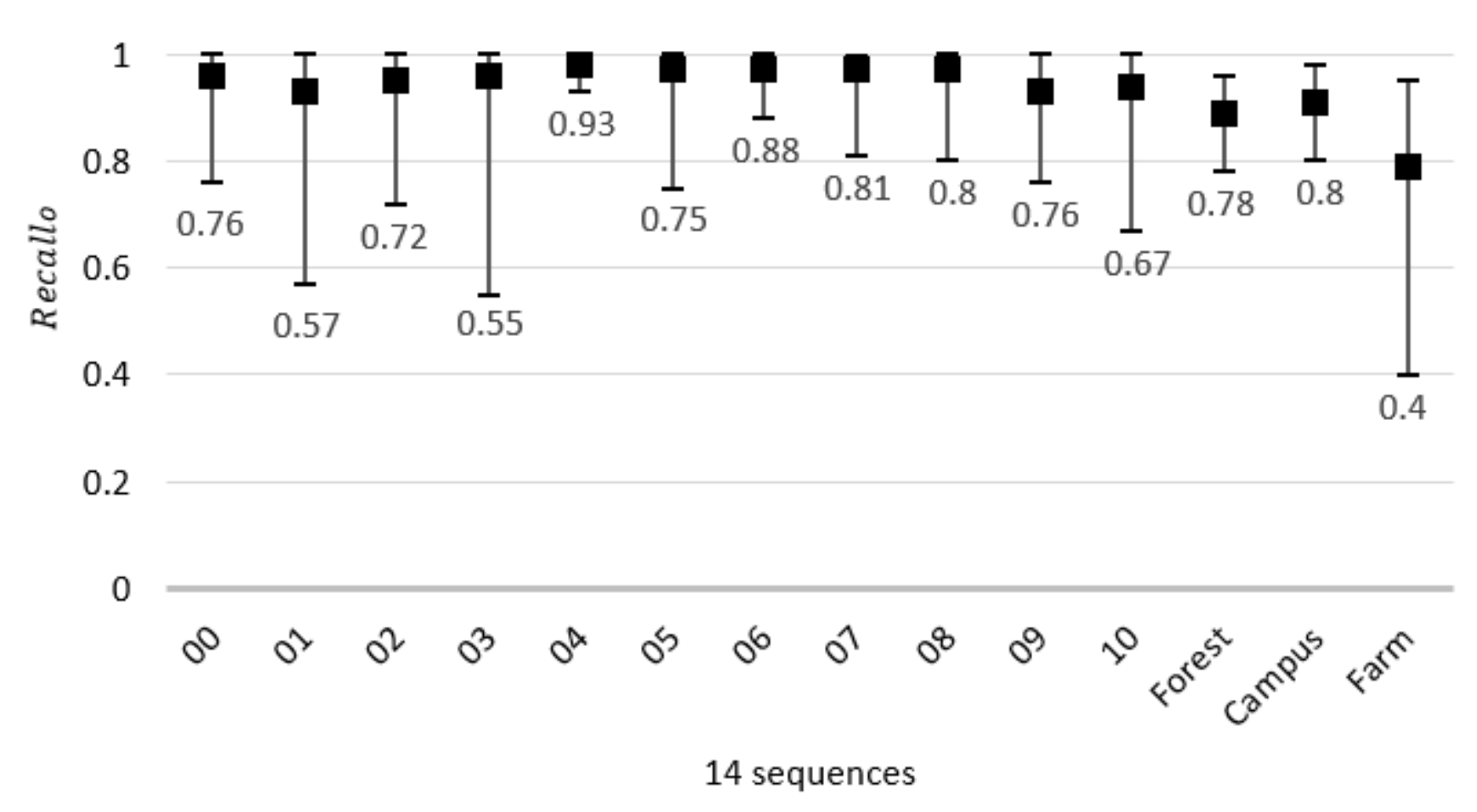}
\caption{The maximum, minimum, and mean values of the $Recall_o$.}
\label{fig_15}
\end{figure}

In an autonomous driving system, the ground segmentation results are the basis for subsequent obstacle clustering and classification. Therefore, it must be ensured that even in the worst-case scenario, sufficient obstacle information can be obtained for subsequent processing. In order to obtain the stability of the proposed algorithm in different scenarios, we analyzed all the ground segmentation results in 14 sequences, and obtained the maximum, minimum, and mean values of the recall rate indicator of “key obstacles,” as shown in Fig. \ref{fig_15}.

\begin{figure}[htbp]
\centering
\includegraphics[width=3.5in]{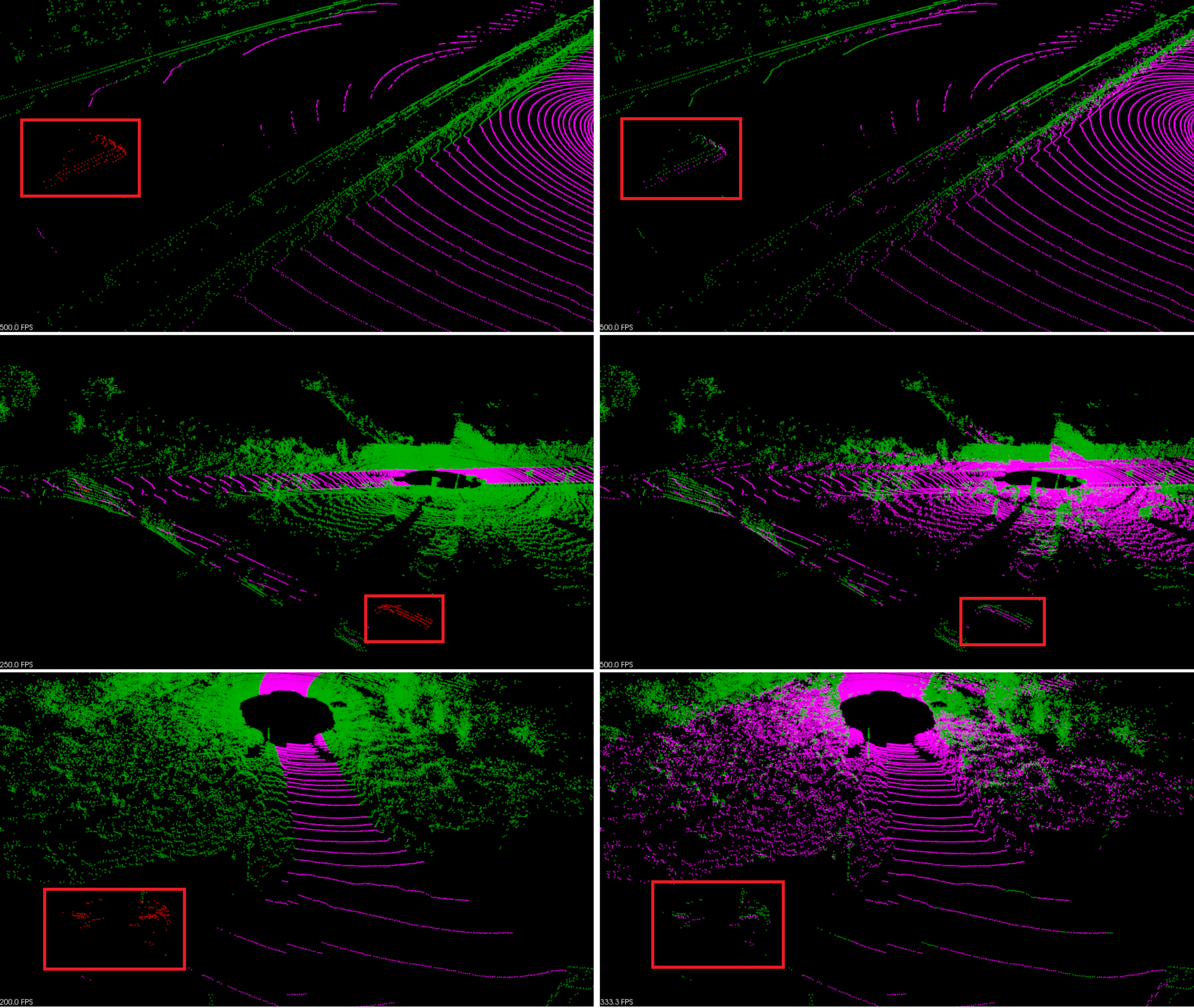}
\caption{Lines 1–3 correspond to the point cloud with the lowest $Recall_o$ in the three sequences “01”, “03”, and “10”. The left column is the benchmark for point cloud classification, and the right column shows the segmentation results.}
\label{fig_16}
\end{figure}

\begin{figure}
\centering
\subfigure[]{\label{fig_17_a}
\includegraphics[width=0.98\linewidth]{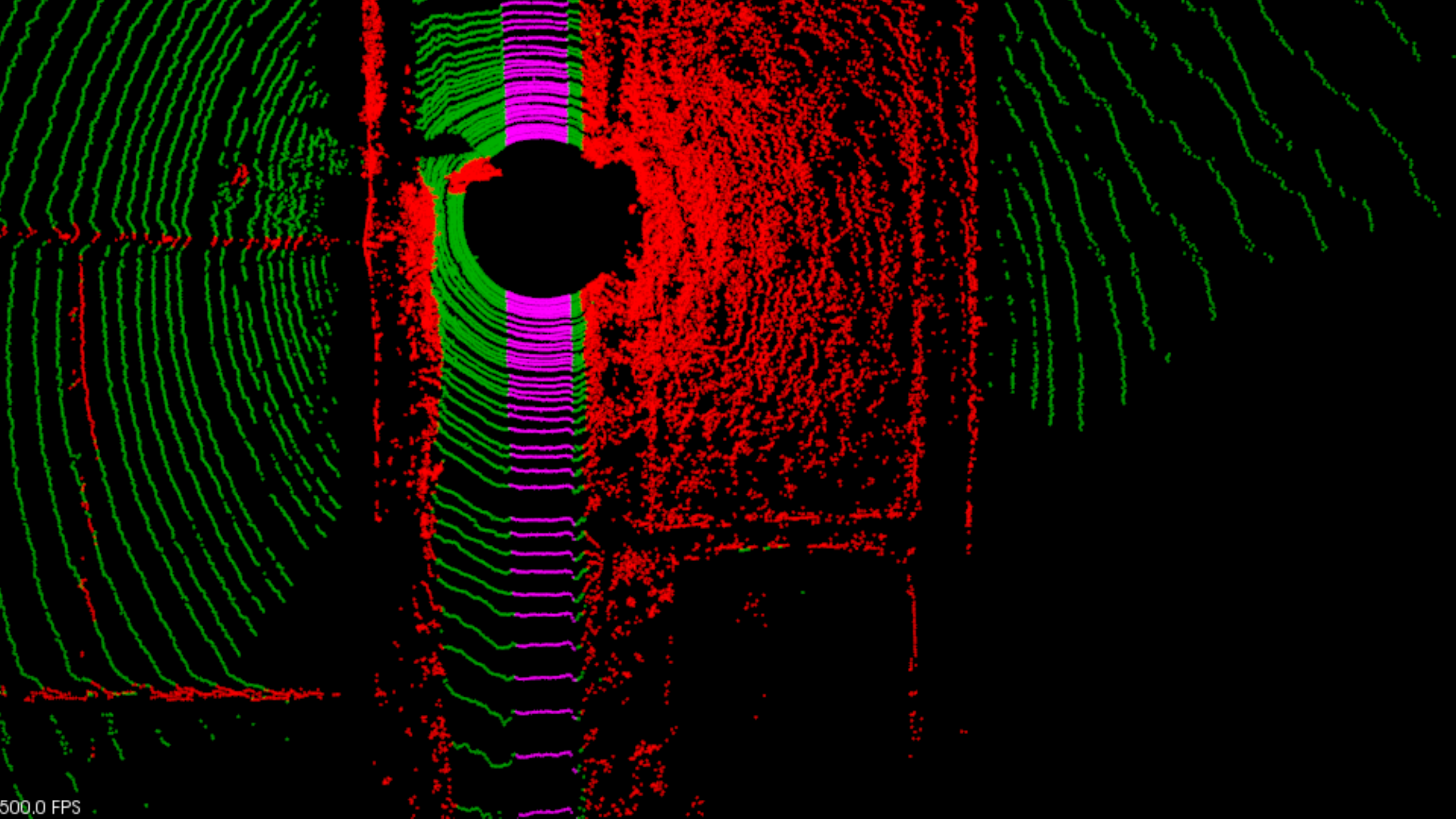}}
\hspace{0.01\linewidth}
\subfigure[]{\label{fig_17_b}
\includegraphics[width=0.98\linewidth]{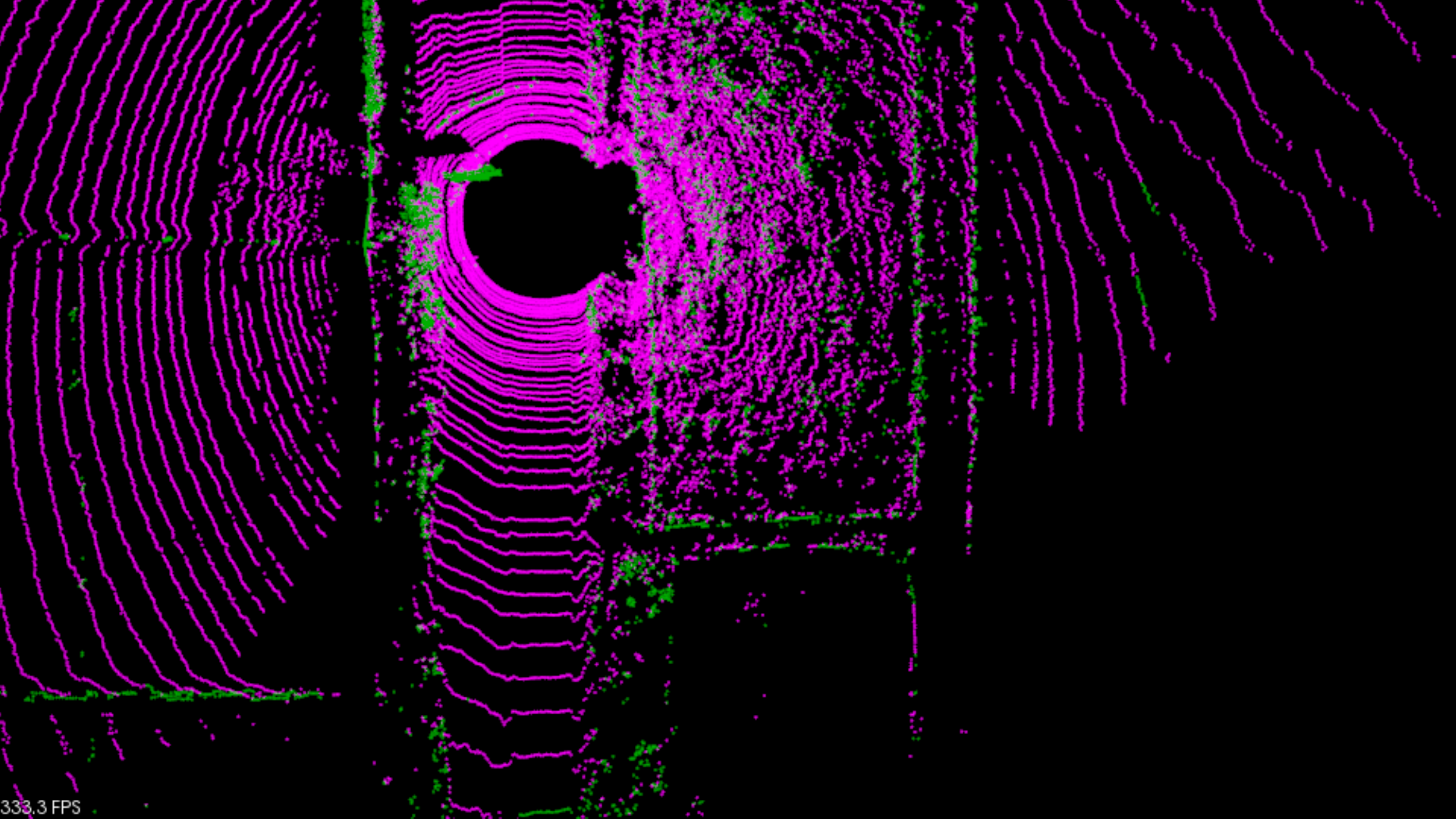}}
\hspace{0.01\linewidth}
\subfigure[]{\label{fig_17_c}
\includegraphics[width=0.98\linewidth]{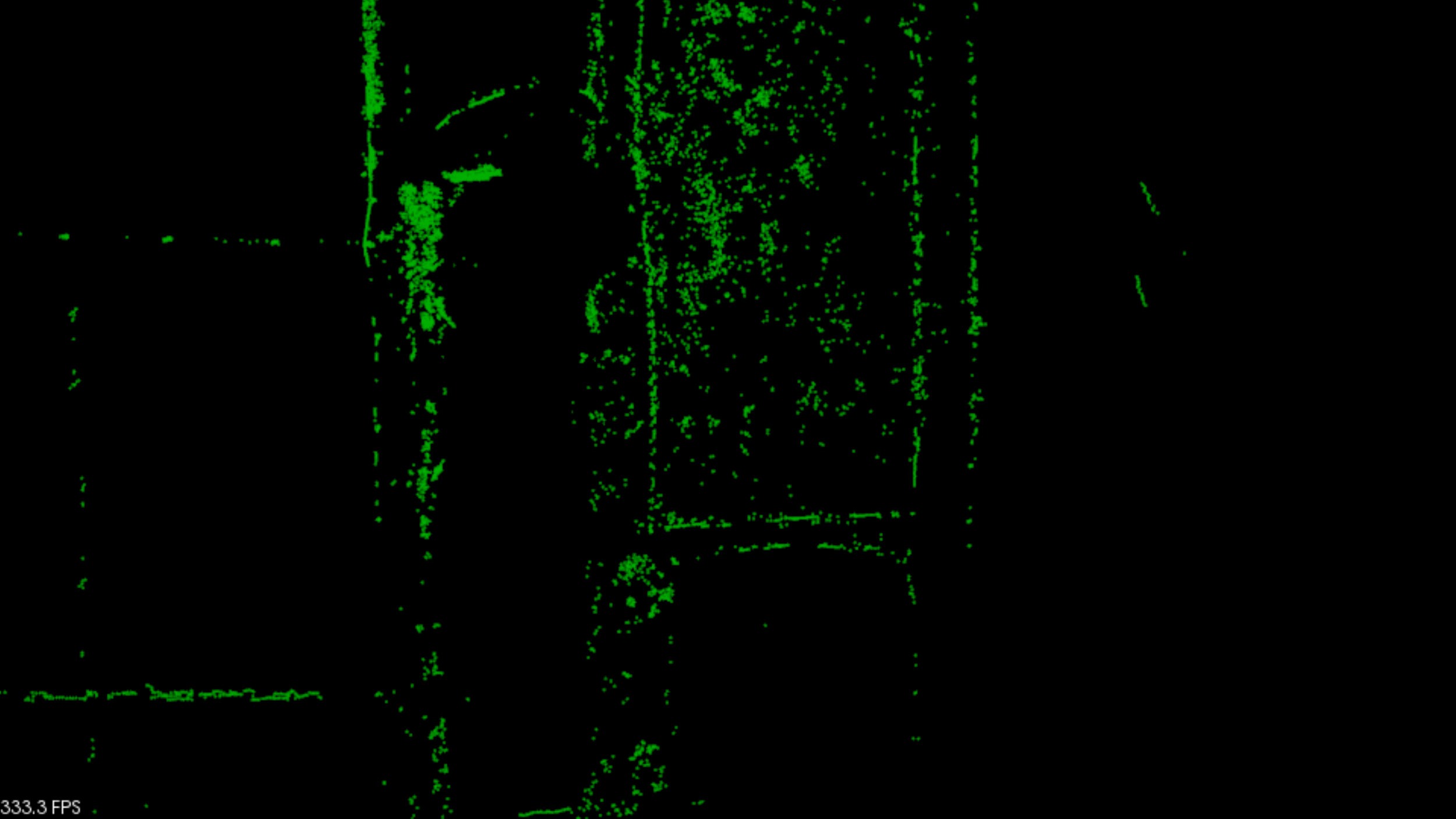}}
\caption{Point cloud with the lowest $Recall_o$ in KoblenzFarmMap: (a) The benchmark for point cloud classification. (b) Segmentation results. (C) Segmentation results without ground point.}
\label{fig_17}
\end{figure}

As can be seen from Fig. \ref{fig_15}, in most cases the proposed algorithm provided good ground segmentation due to the retention of most of the “key obstacle” points. Since the stability of the system is affected by the worst performance, we chose the “01”, “03”, “10”, and “Farm” sequences for the minimum $Recall_o$ of less than 70\% for analysis. Their benchmark for point cloud classification and the segmentation results are shown in Figs. \ref{fig_16} and \ref{fig_17}.

Sequence "01" was a highway scene. As can be seen from the first line of Fig. \ref{fig_16}, the “key obstacle” was in the side lane and was separated by a guardrail. Due to occlusion, there were no ground points around the “key obstacle” for reference. Therefore, the points at the bottom of the vehicle are mistakenly segmented into ground points. Sequence "02" was a residential area and sequence "10" was city traffic—the reasons for their low $Recall_o$ being the same as for "01". However, as can be seen from the right column of Fig. \ref{fig_16}, the segmentation of the obstacle main body was correct, and the segmentation result was sufficient to ensure that the automatic driving system detected the obstacle.

In the KoblenzFarmMap, since vehicles are not allowed to drive into areas such as farmland, we label these areas as “key obstacle.” However, in point cloud data, these regions are similar to the ground, so the algorithm proposed in this paper incorrectly segments the flatter of these objects into ground, resulting in a lower $Recall_o$. Nonetheless, as can be seen in Fig. \ref{fig_17}c, the ground segmentation retained the road boundary information well enough for the autopilot system to travel safely in the area.

\begin{figure}[htbp]
\centering
\includegraphics[width=3.5in]{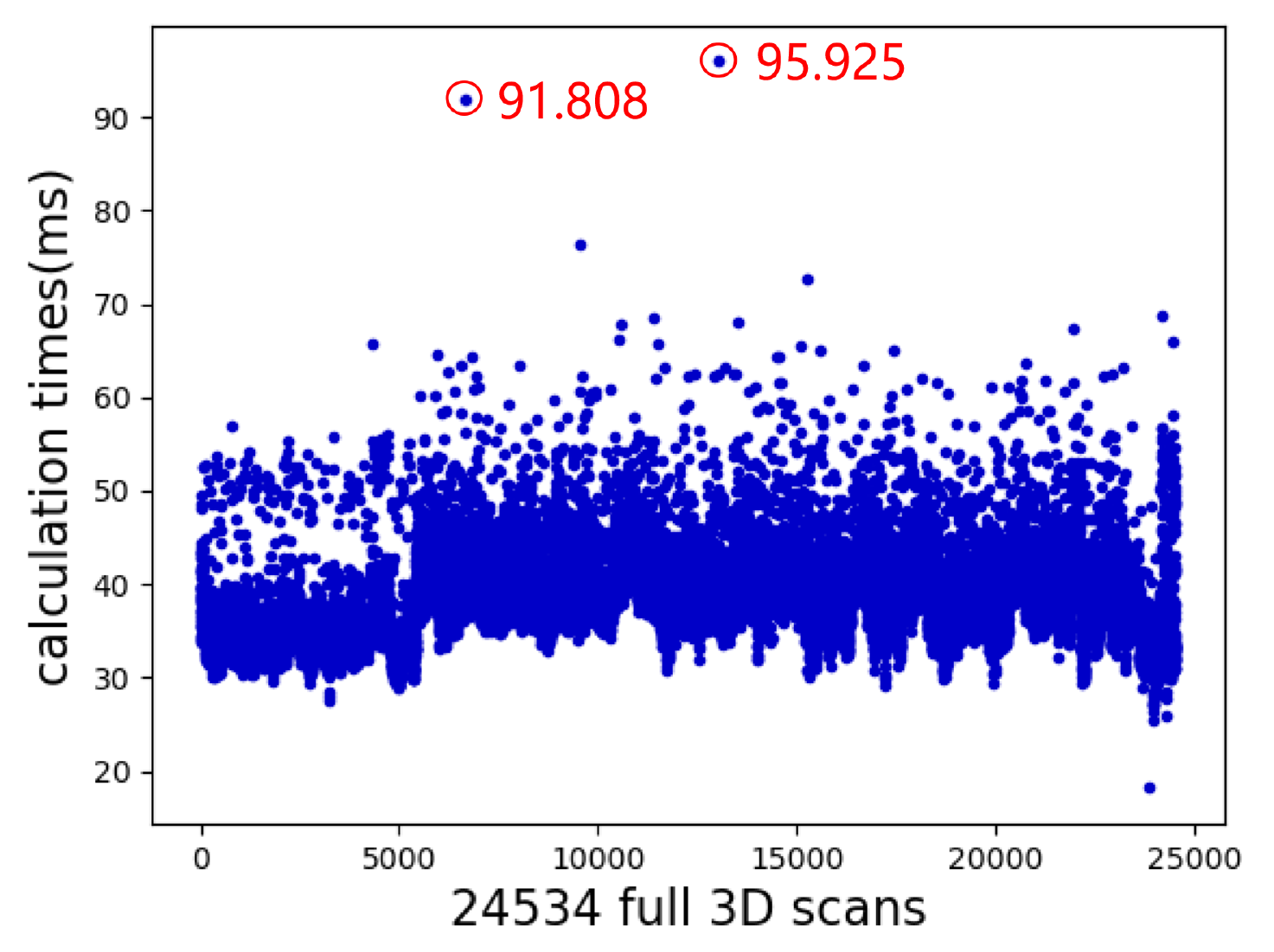}
\caption{The calculation time of our method in 24,534 full 3D scans.}
\label{fig_18}
\end{figure}

In addition to stability, autonomous driving systems also need to ensure that the methods used run in real time. Calculated at 10 HZ point cloud data update speed, the point cloud processing method needs to complete one frame of point cloud processing within 100 ms. In order to test the real-time processing ability of the proposed method, we recorded the calculation time of the algorithm in 24,534 full 3D scans, as shown in Fig. \ref{fig_18}.

It can be seen from Fig. \ref{fig_18} that even in the worst case scenario, the algorithm only needed 95.925 ms to complete the calculation of one frame of point cloud data, which met the requirements of real-time operation.

\begin{figure}[htbp]
\centering
\includegraphics[width=3.5in]{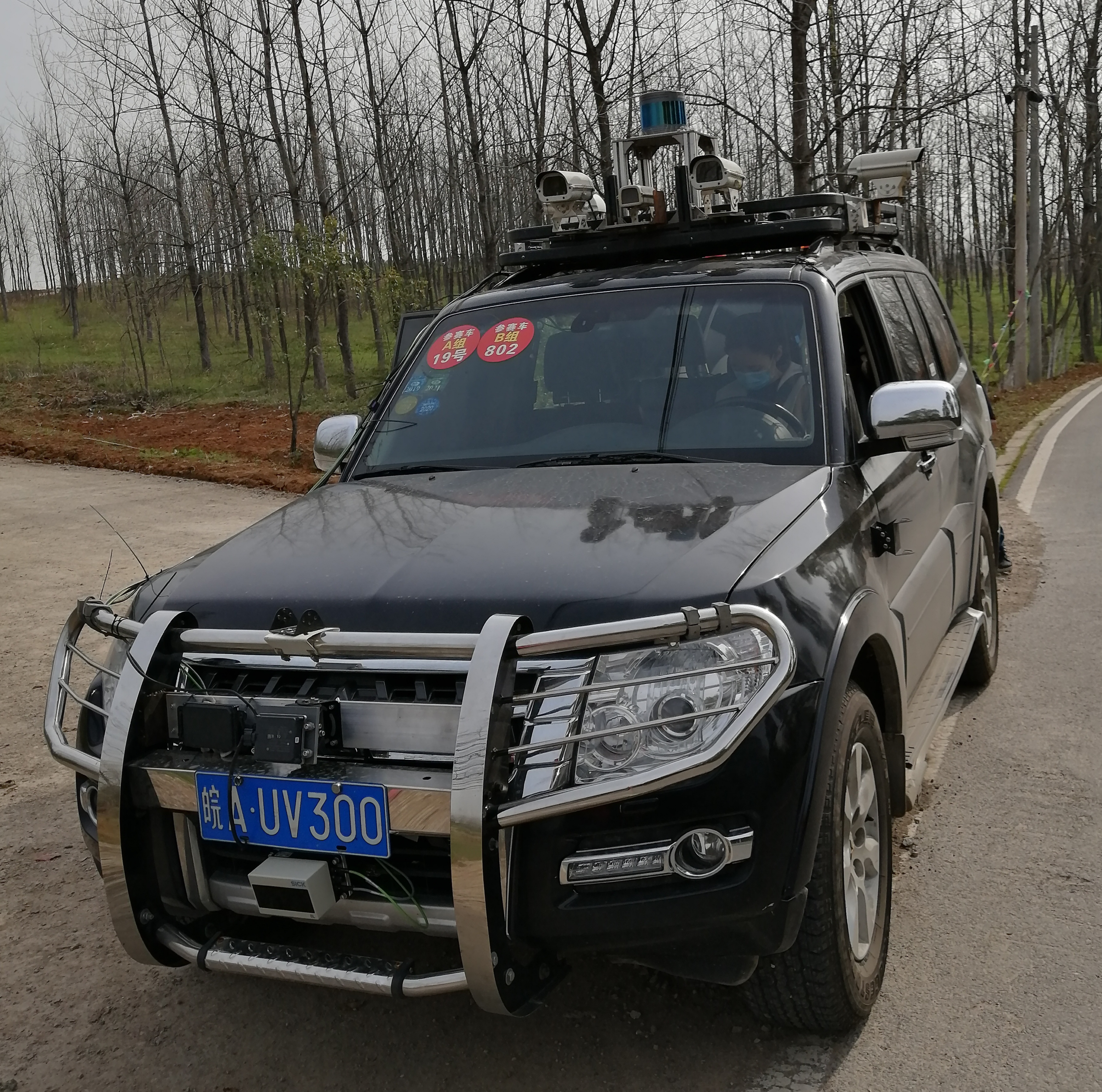}
\caption{ALV used for the experiments in various environments. It is equipped with cameras, lasers, and positioning sensors for perception.}
\label{fig_19}
\end{figure}

To further verify the utility of the algorithm proposed in this paper, we conducted real-vehicle experiments on a self-developed autopilot platform. As shown in Fig. \ref{fig_19}, the platform was a modified Mitsubishi Pajero with an RS-Ruby128 as the main LiDAR. The LiDAR was equipped with 128 lasers that rotated at a speed of 10 Hz and output twice as many points per second as the Velodyne HDL-64E.

\begin{figure}
\centering
\subfigure[]{\label{fig_20_a}
\includegraphics[width=0.45\linewidth]{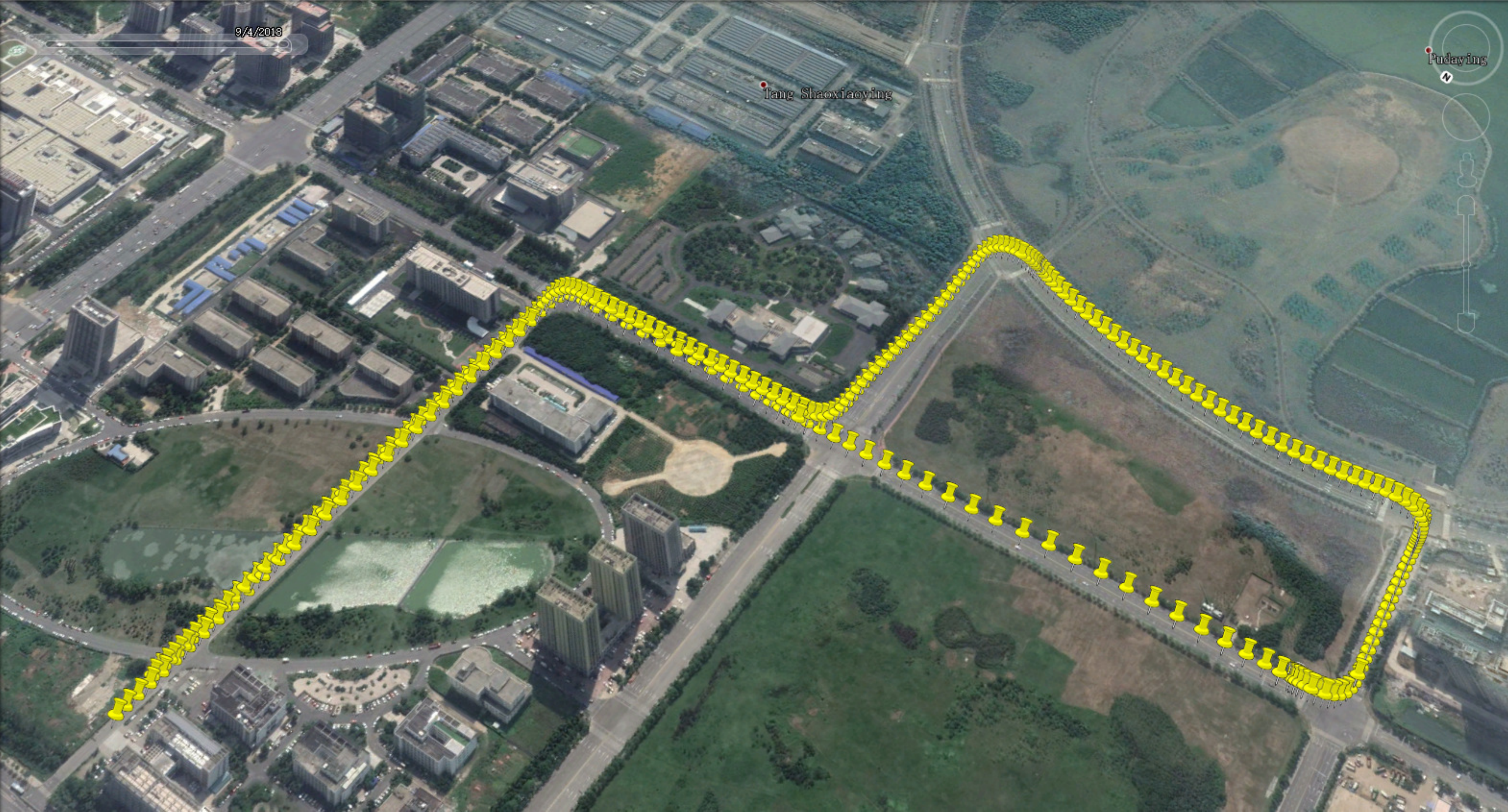}}
\hspace{0.01\linewidth}
\subfigure[]{\label{fig_20_b}
\includegraphics[width=0.45\linewidth]{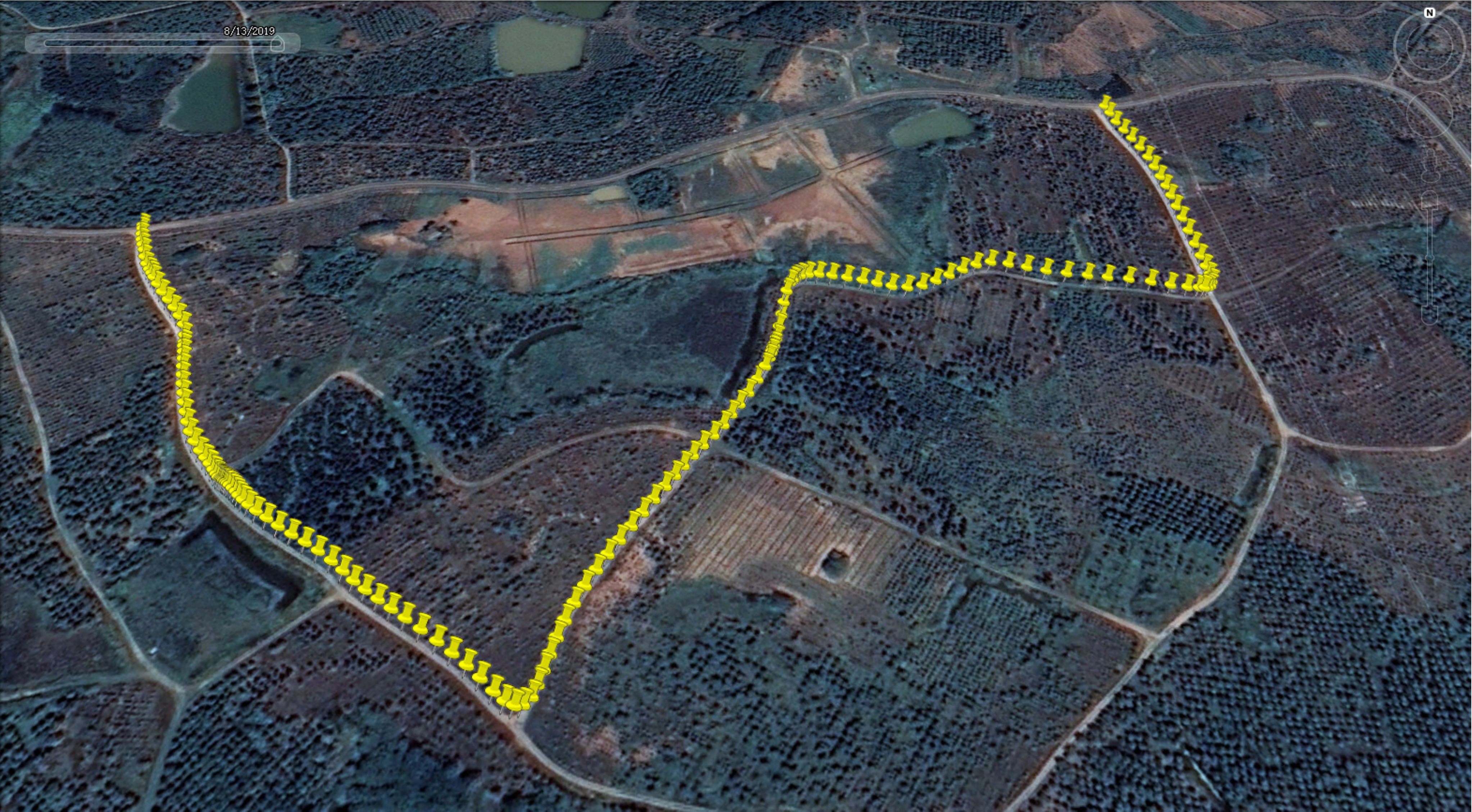}}
\caption{Roadmap for the experiment: (a) urban scene. (b) off-road scene.}
\label{fig_20}
\end{figure}

We experimented with the algorithm in the urban and off-road scenarios of Hefei City. The route of the experiment is shown in Fig. \ref{fig_20}. The urban scene contained 4,307 full 3D scans, while the field scene contained 3,197. The experiments covered the real-time performance of the ground segmentation algorithm and the availability and security of its segmentation results.

\begin{figure}[htbp]
\centering
\includegraphics[width=3.5in]{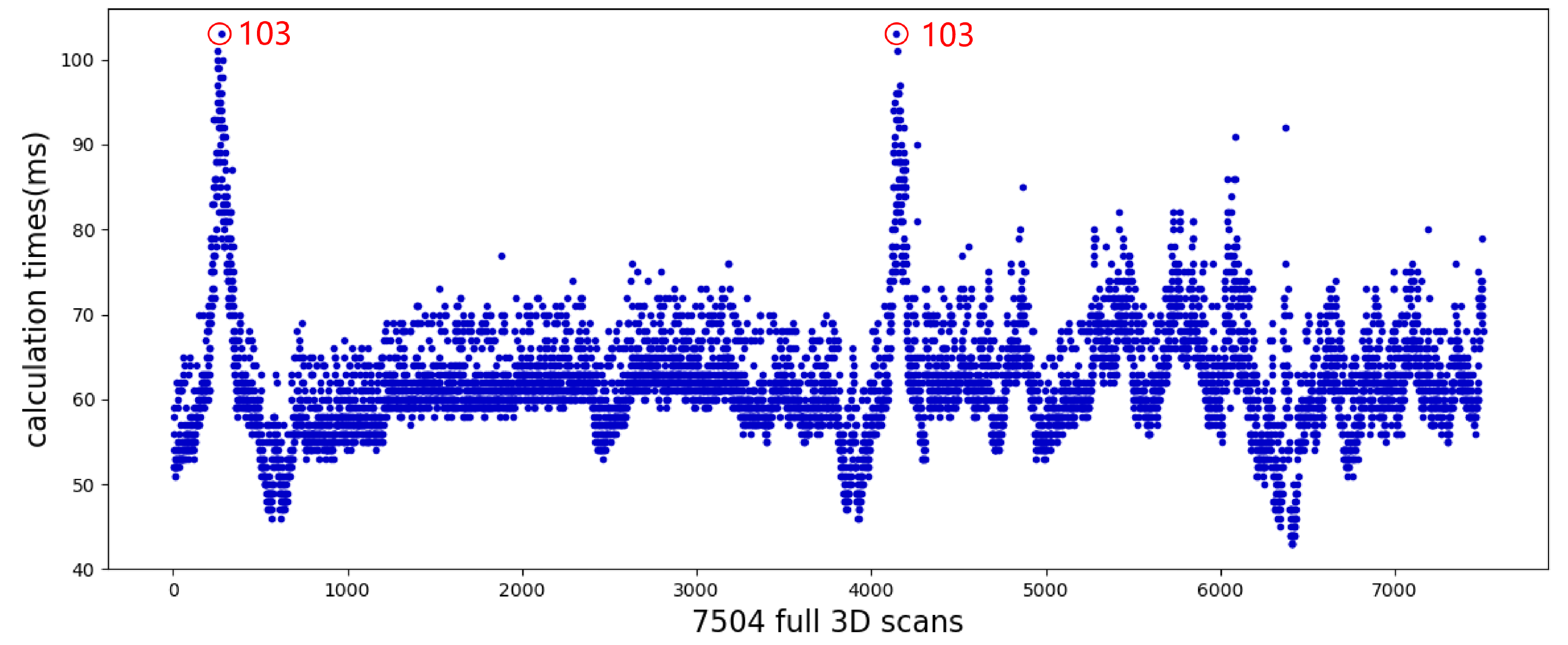}
\caption{Calculation time of our method in 7,504 full 3D scans of real-vehicle experiments.}
\label{fig_21}
\end{figure}

In terms of real time, since the data volume of the RS-Ruby128 is nearly twice that of the Velodyne HDL-64E, it was found after the tests that ground segmentation using the algorithm proposed in this paper required at least 70 ms of processing time, at worst 200 ms, and could not be run in real time. In order to speed up the program, we only performed MRF calculations of the data of 70 lasers that scanned to the ground. As shown in Fig. \ref{fig_21}, the modified algorithm took 103 ms to complete the calculation of one frame of the point clouds in the worst case scenario. Real-time requirements can be met in the vast majority of cases.

\begin{figure}[htbp]
\centering
\includegraphics[width=3.5in]{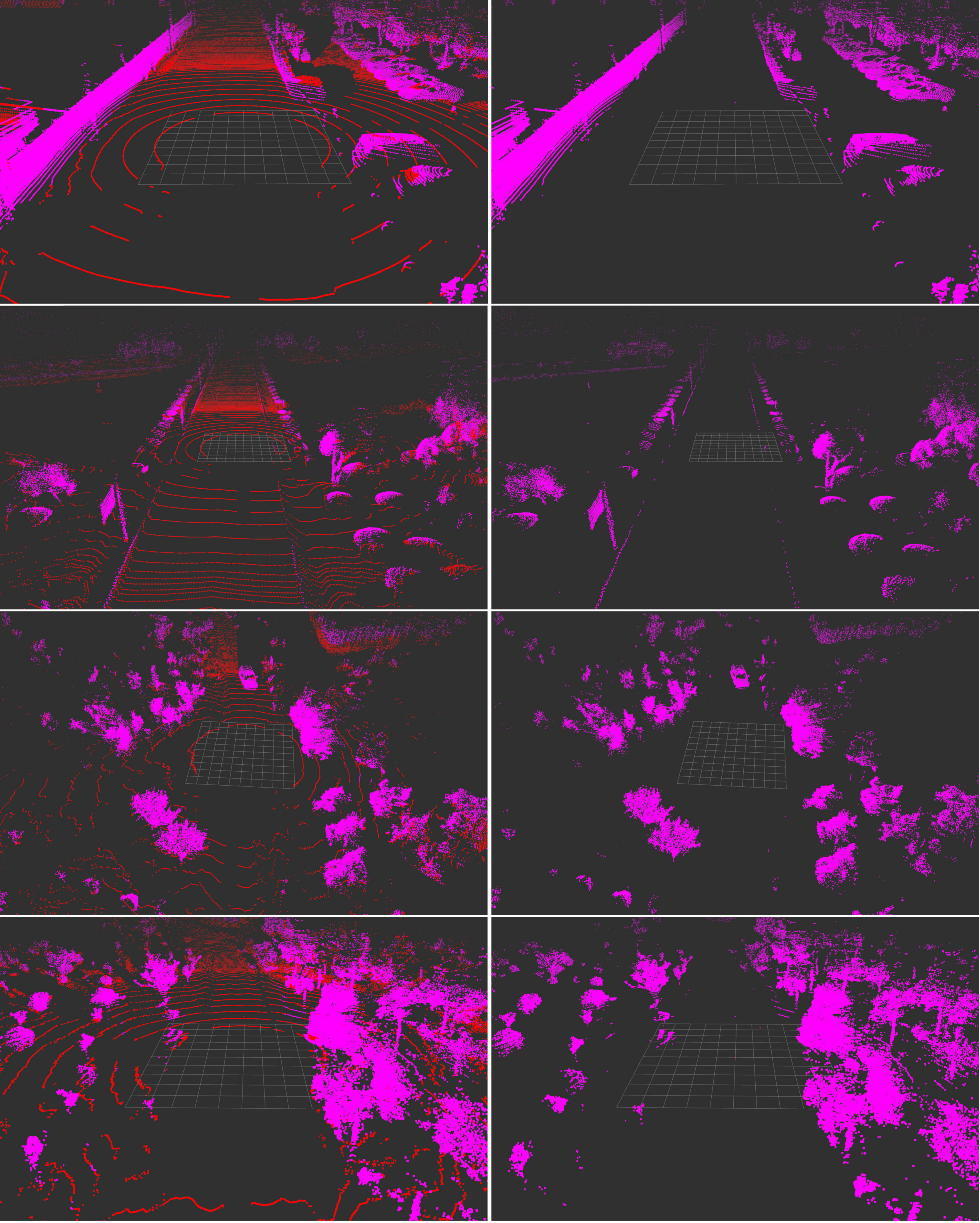}
\caption{Results of the ground segmentation of urban (first two lines) and field (last two lines) scenes. The same line of images corresponds to the same scene. The red points in the left row of images are ground points, while the right row of images shows only obstacle points.}
\label{fig_22}
\end{figure}

In terms of algorithm availability and security, the results of the real-vehicle experiments were consistent with the results in the data set. In urban scenarios, the algorithm was available because it divided the ground well, and there were very few under-segmentation problems. It retained information about obstacles well, so that it could detect some low road edges, thus ensuring security. In field scenarios, the algorithm also had very few segmentation problems and could detect weeds and low undergrowth very well, allowing the autopilot system to obtain sufficient information to keep the vehicle on the road. Some of the experimental results are shown in Fig. \ref{fig_22}.

\section{CONCLUSION AND FUTURE WORK}

In this study, we propose a fast point cloud ground segmentation approach based on coarse-to-fine MRF. The method uses an improved elevation map for ground coarse segmentation, and then uses spatiotemporal adjacent points to optimize the segmentation results. The processed point cloud is classified as high-confidence obstacle points, ground points, and unknown classification points to initialize an MRF model. Then, the graph cut method is used to achieve fine segmentation. Experiments with a total of 24,534 full 3D scans verified the quality of the proposed algorithm in the IOU of ground segmentation and the recall rate of “key obstacles.” Compared to other graph-based methods, the algorithm proposed in this paper is highly efficient and can run in real time. Field tests were also conducted to demonstrate the effectiveness of our proposed method.

Considering that the clustering and recognition of point clouds is also required after ground segmentation, although the ground segmentation algorithm proposed in this paper only requires an average of 39.77 ms to process a frame of Velodyne HDL-64E data, it still occupies nearly 40\% of the processing time. The issue of real time is even more pronounced in cases where there is a greater amount of data (such as in our field tests). Therefore, we will further optimize the algorithm and explore the method of applying the prior feature graph cut method to the clustering task.

\appendix
The two video clips of the field tests on urban and offroad road were also uploaded to Youtube.

Urban scene: https://youtu.be/jeq1vYCwbjA

Offroad scene: https://youtu.be/p4mRsPwmyG4

\begin{IEEEbiography}[{\includegraphics[width=1in,height=1.25in,clip,keepaspectratio]{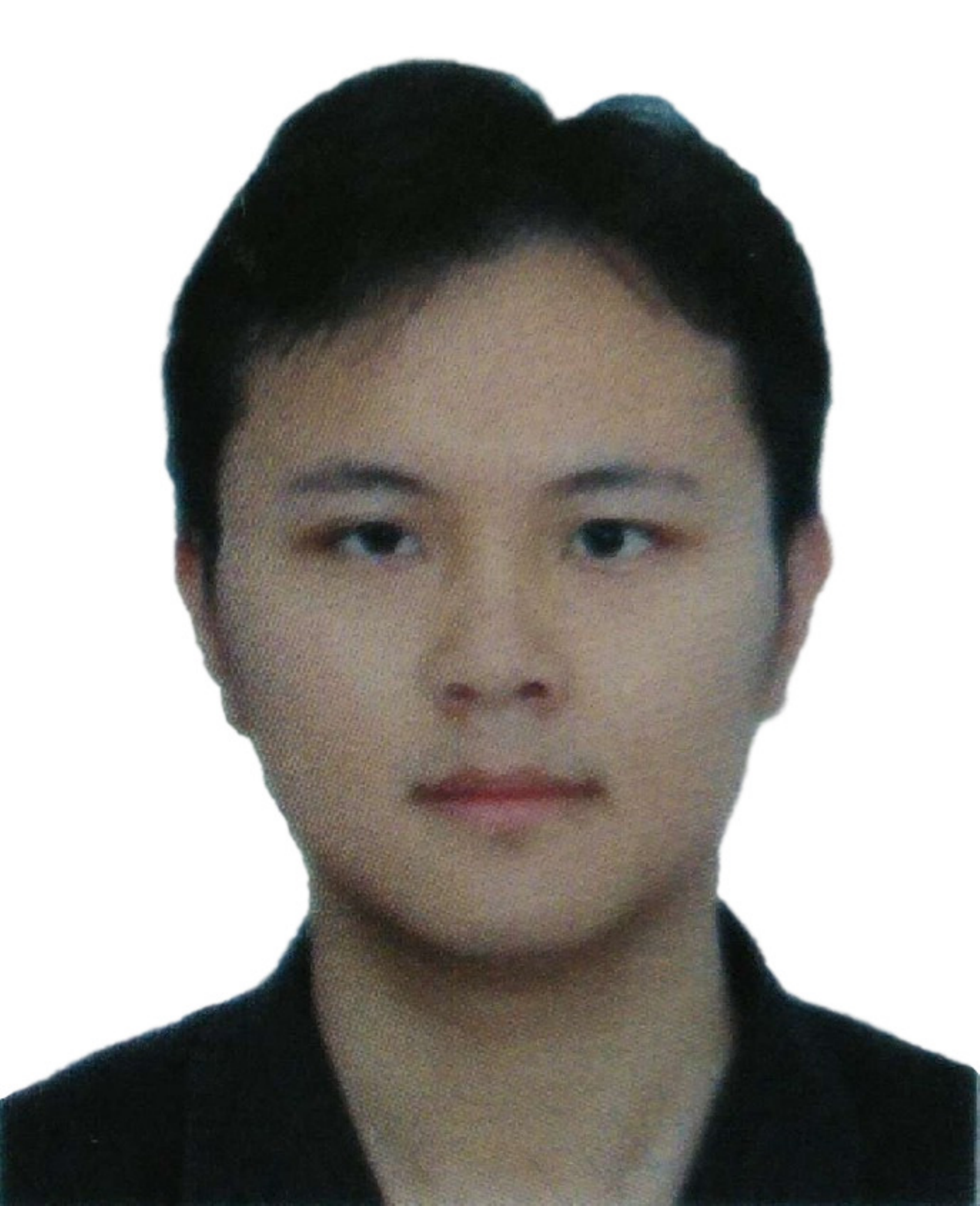}}]{Weixin Huang}
received the M.S. degree in integrated circuits engineering from Shenzhen University, Guangdong, China in 2016. He is currently working toward the Ph.D. degree in detection technology and automation from the University of Science and Technology of China, and his laboratory is at Hefei Institutes of Physical Science, Chinese Academy of Sciences. His research interests involve ground segmentation, 3D object clustering and classification in intelligent transportation systems.
\end{IEEEbiography}

\begin{IEEEbiography}[{\includegraphics[width=1in,height=1.25in,clip,keepaspectratio]{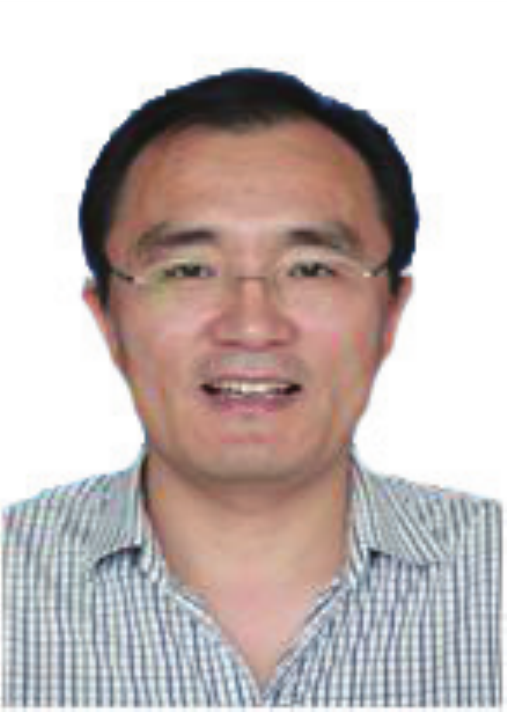}}]{Huawei Liang}
received a Ph.D. degree in detection technology and automation from the University of Science and Technology of China in 2007. Now, He is a principal investigator and the deputy director of the Institute of Intelligent Machines, Hefei Institutes of Physical Science Chinese Academy of Science. He has been engaged in the robotics, intelligent vehicle technology and systems, detection technology and automation device, pattern recognition and intelligent system, control theory and control engineering.
\end{IEEEbiography}

\begin{IEEEbiography}[{\includegraphics[width=1in,height=1.25in,clip,keepaspectratio]{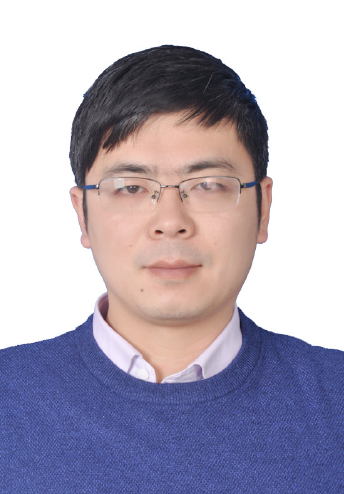}}]{Linglong Lin} 
received the B.S. degree in computer science and technology from Anhui University of Technology in 2010, the Ph.D. degree in nuclear science and engineering from the University of Chinese Academy of Sciences in 2016. He is currently an associate professor at the Institute of Intelligent Machines, Hefei Institutes of Physical Science, C.A.S, Hefei, China. Since 2016. His research has focused on self-driving vehicles, such as LiDAR point cloud data processing, object recognition and tracking, and deep learning.
\end{IEEEbiography}

\begin{IEEEbiography}[{\includegraphics[width=1in,height=1.25in,clip,keepaspectratio]{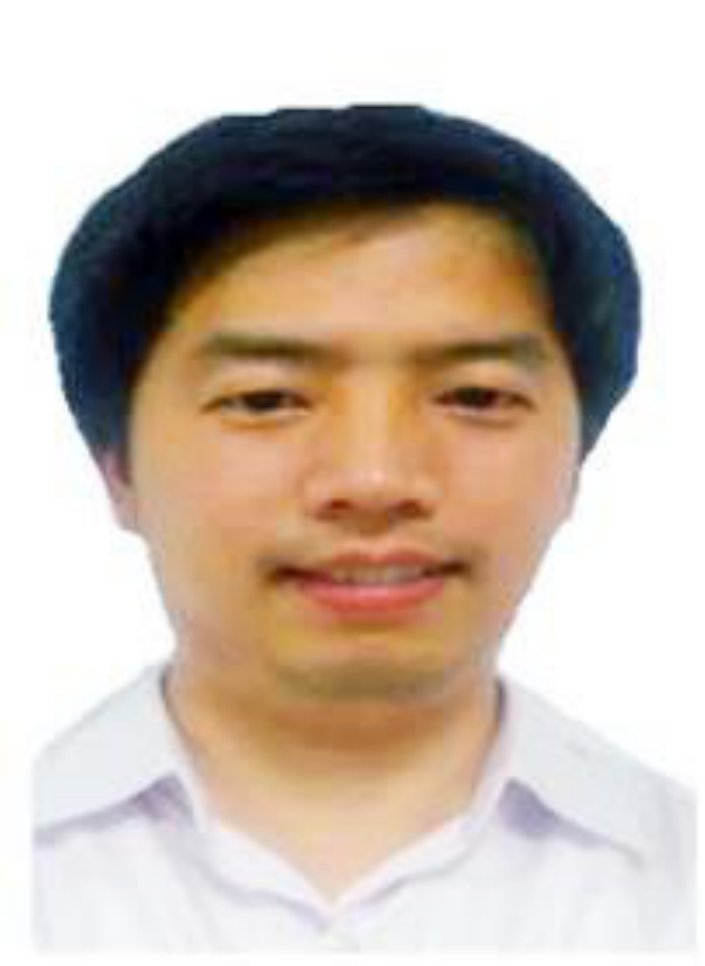}}]{Zhiling Wang}
received a Ph.D. degree in control science and engineering from the University of Science and Technology of China in 2008. Now, he is a master tutor in the University of Science and Technology of China, Institute of Intelligent Machines, Hefei Institutes of Physical Science Chinese Academy of Science. He has been mainly engaged in the research of driverless vehicles, environmental perception and understanding, machine vision and machine learning.
\end{IEEEbiography}

\begin{IEEEbiography}[{\includegraphics[width=1in,height=1.25in,clip,keepaspectratio]{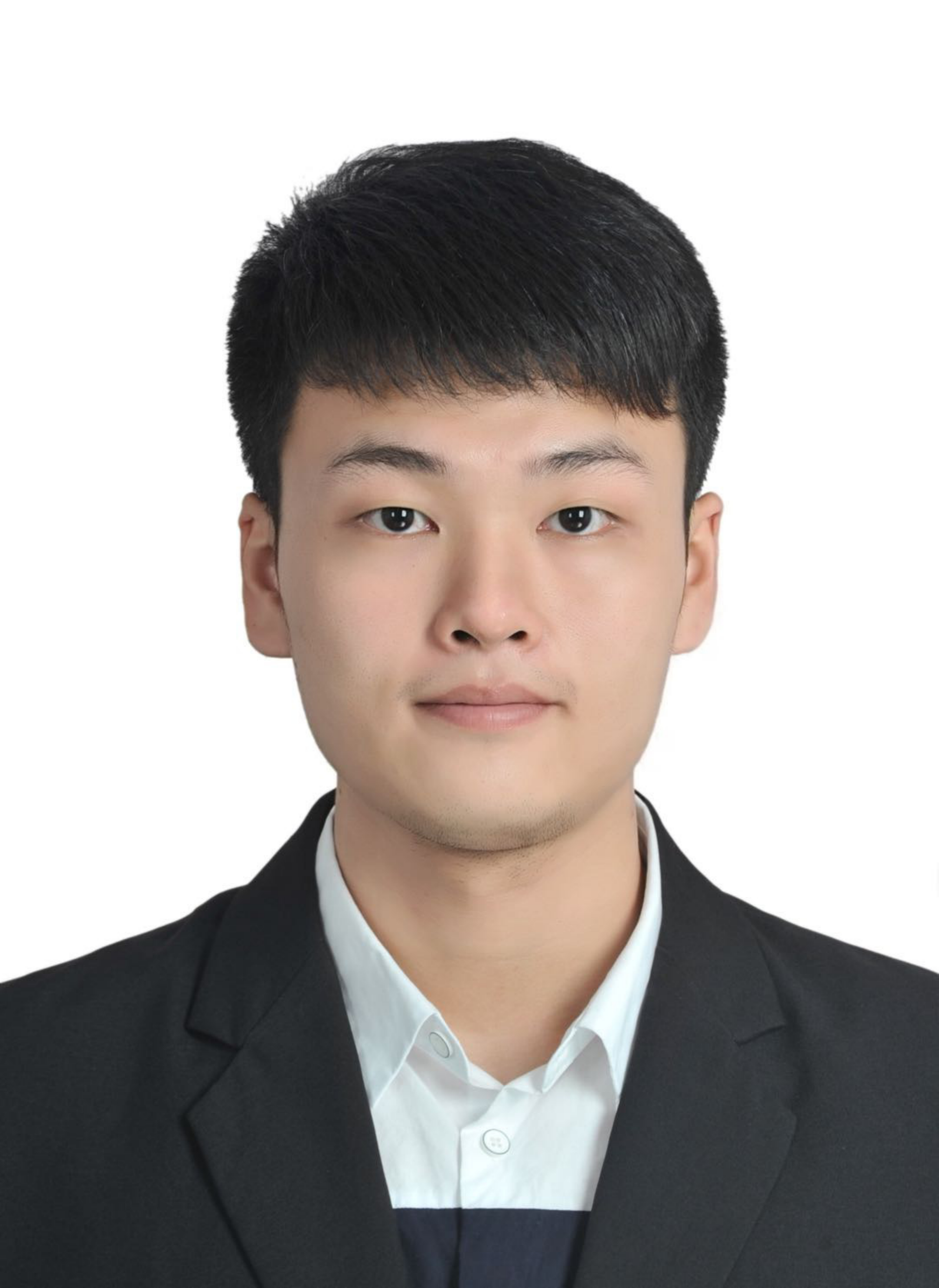}}]{Saobo Wang}
received the B.S. degree in software engineering from Heilongjiang University, Harbin, China, and the M.S. degree in integrated circuits engineering from Shenzhen University, Guangdong, China, in 2014 and 2017, respectively. He is currently a Ph.D. student in Pattern Recognition and Intelligent Systems at University of Science and Technology of China, Anhui, China. His research interets include motion planning, maneuver prediction and driving strategies for highly automated driving.
\end{IEEEbiography}

\begin{IEEEbiography}[{\includegraphics[width=1in,height=1.25in,clip,keepaspectratio]{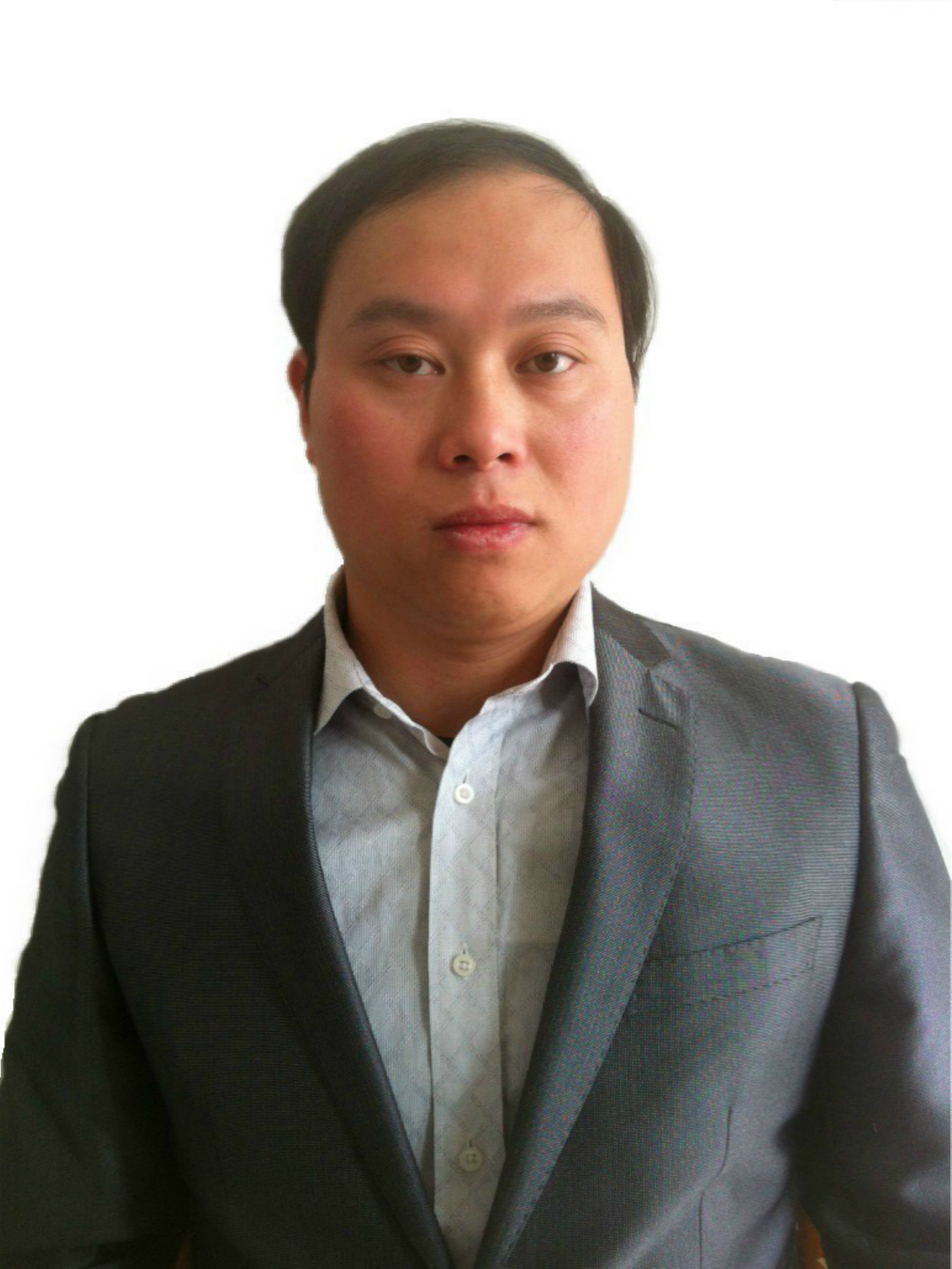}}]{Biao Yu}
received his Ph.D. and M.S. degrees from School of Mechanical and Automotive Engineering, Hefei University of Technology in 2013 and 2010 respectively, and his B.S. degree from School of Mechanical and Automotive Engineering, Hefei University of Technology in 2007. He is an associate professor at the Institute of Intelligent Machines, Hefei Institutes of Physical Science, Chinese Academy of Sciences, China. His research interests include evolutionary computation, intelligent vehicle and mobile robot navigation and localization. 
\end{IEEEbiography}

\begin{IEEEbiography}[{\includegraphics[width=1in,height=1.25in,clip,keepaspectratio]{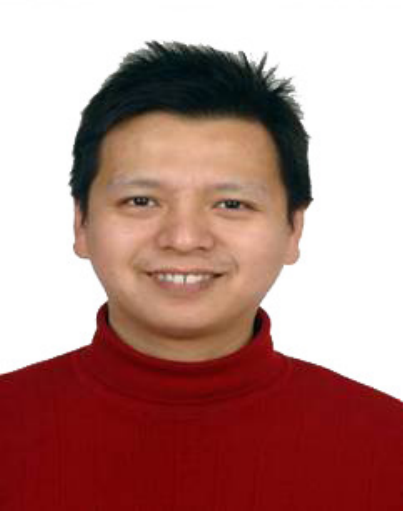}}]{Runxin Niu}
received a Ph.D. degree from the Jiangsu University in 2007. He is a researcher of the Hefei Research Institute of the Chinese Academy of Sciences, a doctoral tutor at the University of Science and Technology of China, an expert in the preparation of the special implementation plan for the “smart agricultural machinery equipment” of the national key R\&D plan. His research interest is smart agriculture equipment.
\end{IEEEbiography}

\end{document}